\documentclass{article}
\usepackage{iclr2027_preprint,times} 

\usepackage{microtype}
\usepackage{graphicx}
\usepackage{subcaption}
\usepackage{booktabs}
\usepackage{multirow}
\usepackage{amsmath,amssymb,mathtools,amsthm}
\usepackage{xcolor}
\usepackage{algorithm}
\usepackage{algpseudocode}
\usepackage{url}
\usepackage{tikz}
\usetikzlibrary{arrows.meta,positioning,fit}
\usepackage{hyperref}
\hypersetup{colorlinks=false,pdfborder={0 0 0}}
\usepackage[capitalize,noabbrev]{cleveref}

\newcommand{\SORDvisiblebox}[2]{%
  \begingroup
    \setlength{\fboxrule}{0.35pt}%
    \setlength{\fboxsep}{0.30pt}%
    \fcolorbox{#1}{white}{#2}%
  \endgroup
}
\let\SORDorigcitep\citep
\let\SORDorigcitet\citet
\let\SORDorigCref\Cref
\let\SORDorigcref\cref
\RenewDocumentCommand{\citep}{m}{\SORDvisiblebox{green}{\SORDorigcitep{#1}}}
\RenewDocumentCommand{\citet}{m}{\SORDvisiblebox{green}{\SORDorigcitet{#1}}}
\RenewDocumentCommand{\Cref}{m}{\SORDvisiblebox{red}{\SORDorigCref{#1}}}
\RenewDocumentCommand{\cref}{m}{\SORDvisiblebox{red}{\SORDorigcref{#1}}}
\newcommand{\SORDboxedhref}[2]{\SORDvisiblebox{cyan}{\href{#1}{#2}}}

\newcommand{\method}{\ensuremath{\mathrm{SORD}}}

\theoremstyle{plain}

\theoremstyle{definition}

\theoremstyle{remark}

\title{Support-guided OOD Removal with Diffusion}

\author{
  Alireza Kheirandish \& Sara Fridovich-Keil$^{1}$ \\
  $^{1}$Georgia Institute of Technology, Atlanta, GA, USA \\
  \texttt{\{akheirandish3,sfk\}@gatech.edu}
}

\begin{document}
\maketitle
\vspace{-10pt}
\begin{abstract}

Many real-world signals are corrupted by localized, structured perturbations whose distribution is unknown: an intermittent background noise in a music recording, a watermark stamped on a photograph, a cast shadow. We study the task of recovering a clean signal $x$ from an observation $y$ containing such a localized out-of-distribution (OOD) corruption. 
We propose Support-guided OOD Removal with Diffusion, $\method{}$, an iterative algorithm to recover $x$ using only two ingredients: a binary map of the corruption support and a single diffusion model pretrained on clean signals. 
$\method{}$ does not require any paired data or any statistical model of the corruption. 
We evaluate $\method{}$ on natural images and human faces with spatially localized corruptions including handwritten digits, other images, watermarks, and shadows, as well as audio recordings of different musical instruments with temporally localized corruptions including other instruments and human speech.
$\method{}$ is competitive with, and often outperforms, posterior-sampling inpainting given the same diffusion prior and the same mask, source-separation baselines that require a separate diffusion model for each source, and, on several benchmarks, supervised networks trained on paired clean and corrupted signals. The code is released here: \SORDboxedhref{https://github.com/voilalab/SORD}{https://github.com/voilalab/SORD}.
\end{abstract}

\section{Introduction}
\label{sec:intro}

\begin{minipage}{\linewidth}
    \centering
    \includegraphics[
        width=\linewidth
    ]{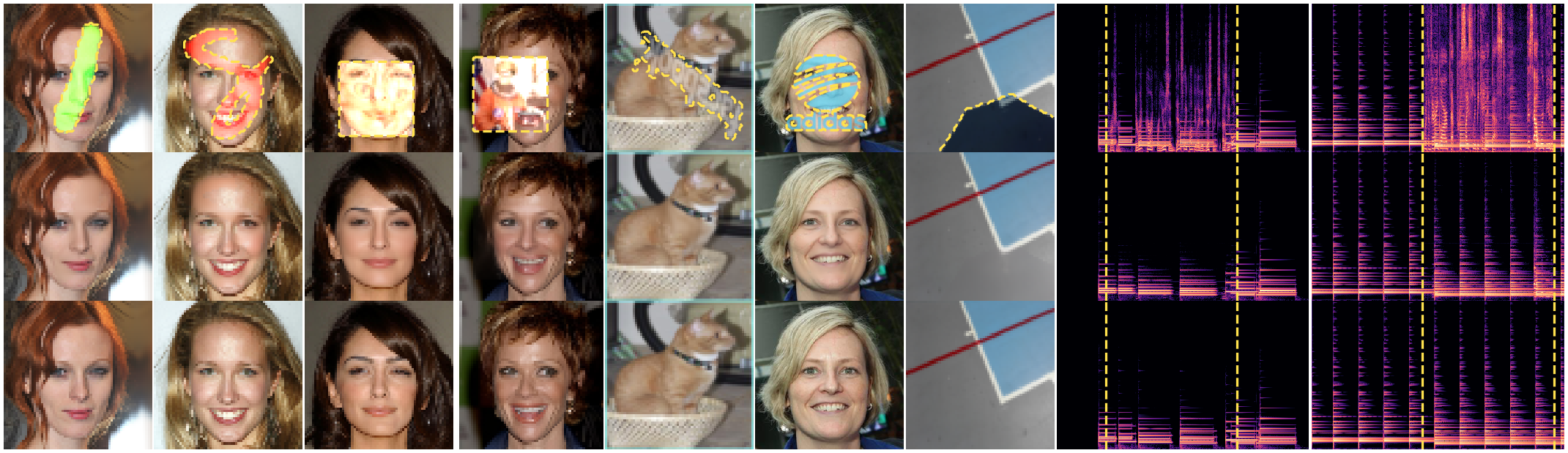}

    \captionof{figure}{
        \textbf{$\method{}$ removes diverse localized OOD corruptions} without prior knowledge of the corruption distribution.
        Examples span semantic image patches, watermarks, shadows, and
        localized audio interference. Rows show the corrupted observation (top),
        our reconstruction (middle), and the
        clean target (bottom).
    }
    \label{fig:teaser}
\end{minipage}

Localized corruptions arise in many restoration problems: a limited region of
an image may be affected by structured noise, a watermark may cover part
of an image, a cast shadow may alter a region, or an interfering
sound may overlap a musical recording for a short time. These corruptions can be
highly structured and semantically meaningful, rendering simple (e.g., Gaussian) noise models
inadequate. At the same time, collecting paired examples or training a
generative model for every possible corruption type can be impractical in the face of unexpected corruptions.

Existing approaches for removing localized corruptions generally fall into
three categories. Supervised restoration methods learn a direct mapping
from corrupted to clean signals~\citep{liu2021wdnet,liang2021slbr,guo2023shadowdiffusion}; these methods can perform well on specific tasks, 
but require paired clean and corrupted training data and are then restricted to removing corruptions similar to those seen during training.
Posterior-sampling approaches combine a pretrained generative prior
over clean signals with a known measurement model or likelihood
~\citep{kawar2022ddrm,chung2023dps}.
While these methods have achieved strong performance on many inverse problems~\citep{lugmayr2022repaint,wang2023ddnm,song2023pigd}, including inpainting a masked region, they generally require the corruption statistics to be known \citep{legin2023slic}, often assuming relatively simple likelihood models such as Gaussian or Poisson noise \citep{chung2023dps}
rather than corruptions with complex spatial, semantic, or temporal structure.
Finally, source-separation approaches explicitly learn distributions for both
the clean signal class and the corruption class, for example using separate generative models \citep{stevens2025removing,mariani2024msdm}.
These approaches can represent substantially richer corruption types, but require
training data or learned generative priors for the anticipated corruption classes.

In many applications, these requirements can be restrictive or impractical. Paired clean and corrupted examples may be unavailable,
and the corruption distribution may be too complex or unexpected to model accurately. We therefore consider a more weakly specified setting in which a strong generative prior is available for the clean signal class and the corruption is localized to a known support $\Omega$; we do not assume access to any paired corruption data, parametric corruption distribution, or generative model of the corruption class.

Our key idea is to replace statistical prior knowledge of the corruption distribution with structural prior knowledge of where or when the corruption occurs. In many applications, the corruption is localized: it affects only a spatial subset of an image, as with watermarks, shadows, or localized artifacts, or only a temporal segment of an audio signal, as when an interfering sound briefly overlaps with a recording. The corruption within this region may have complex and unknown structure, but identifying its support is often substantially easier than modeling its distribution. A broad class of anomaly-detection and segmentation
methods have already been proposed to estimate the support of such localized corruptions~{\small\citep{defard2021padim,roth2022patchcore,wyatt2022anoddpm,zhang2023diffad,
kheirandish2026klip}}. 

Rather than attempting to model the corruption distribution, we treat localized corruptions as out-of-distribution (OOD) relative to a generative prior over clean signals.
The observations outside the corrupted region $\Omega$ are regarded as clean and are preserved, while the region inside $\Omega$ is reconstructed using the clean-signal generative prior. Unlike standard inpainting, however, the corrupted observations inside $\Omega$ may still carry information about the underlying clean signal. Our algorithm, $\method$, therefore alternates between explaining the observed corrupted region and reconstructing a signal that is consistent with both the learned clean-data distribution and the trusted observations outside $\Omega$. At each iteration, $\method$ uses its estimated uncertainty to adapt its effective step size and evaluate an adaptive stopping criterion.
Empirically, we find that knowledge of the clean-signal prior and corruption support alone can be sufficient to match or outperform approaches that rely on paired training data or explicit corruption models.

We evaluate $\method$ across localized structured corruptions in both images
and audio, as shown in \Cref{fig:teaser}. For images, we consider both controlled additive corruptions to face images as well as specific applications in watermark removal and shadow removal in which the corruptions are not purely additive. For audio, we consider temporally localized additive audio interference in which a musical recording of one instrument is corrupted by another instrument or by human speech.
Across these settings, we compare $\method$ with posterior-sampling (inpainting) methods
using the same clean-signal prior and corruption support, supervised restoration models trained
for specific corruption types, and multi-source diffusion models that
explicitly model possible interfering sources.
The results show that support information can substantially reduce the need
for an explicit corruption model, while corruption-specific supervision
remains advantageous when the test corruption is well represented by its
training distribution.
To summarize, our main contributions are:
\begin{enumerate}
\item We introduce $\method$, an iterative algorithm for removing localized, highly structured OOD corruptions without requiring prior knowledge, training examples, or a generative model of the corruption. Instead, our method requires only a mask specifying the corruption support and a diffusion model trained on the clean signal distribution.

\item We demonstrate the generality of our approach across both image and audio restoration tasks, where corruptions are localized spatially or temporally. Across a diverse range of structured corruption types, $\method$ is competitive with approaches with substantially stronger modeling or training requirements, including supervised methods that require paired clean and corrupted data and source-separation methods that require an additional generative model trained on the corruption class.
\end{enumerate}

\section{Related Work}
\label{sec:related}

Our work lies at the intersection of diffusion-based inverse problems,
corruption-specific restoration, source separation, and localized
anomaly detection. 

\vspace{-7px}
\paragraph{Diffusion priors for inverse problems.}
Diffusion models have been widely used as learned priors for inverse problems \citep{song2023pigd, chung2023dps, zhu2023diffpir, zhang2025daps}, including inpainting a masked region \citep{lugmayr2022repaint, kawar2022ddrm, wang2023ddnm}. 
These methods typically assume a specified observation or measurement model together with a known or modeled noise distribution such as Gaussian noise, Poisson noise, and other non-Gaussian likelihoods~\citep{chung2023dps,legin2023slic}. 
In contrast, localized structured corruptions need not follow a known parametric noise model or likelihood, motivating restoration under support information alone.
\vspace{-6px}
\paragraph{Corruption-specific image restoration.}
A complementary line of work trains supervised or generative restoration
models for a particular family of corruptions. 
For example, specialized methods have been proposed for image watermark removal \citep{cun2021split, liu2021wdnet, liang2021slbr} and shadow removal \citep{guo2023shadowdiffusion, guo2023shadowformer}, using either classical supervised learning \citep{cun2021split, liu2021wdnet, liang2021slbr}, conditional diffusion \citep{guo2023shadowdiffusion} or an attention-based model
\citep{guo2023shadowformer}. However, these approaches all require paired clean and corrupted training images and are specialized to specific types of corruption, whereas we seek a single general-purpose algorithm that can remove any structured localized corruption without retraining for each corruption class.
\vspace{-7px}

\paragraph{Generative priors for source separation.}
A broad family of methods focuses on source separation, treating a single observation as a mixture of
multiple constituent sources and using learned generative priors to recover the
individual components. 
For images, \citet{stevens2025removing} model an observation as
the combination of a clean image and structured noise, using separate diffusion
priors for the signal and noise distributions and jointly inferring the two
components. For audio, Multi-Source Diffusion Models
(MSDM)~\citep{mariani2024msdm} learns the joint distribution of a fixed
set of musical sources, such as different instruments, and uses the resulting model for both generation and
source separation. Related diffusion-based approaches learn source
distributions for speech~\citep{scheibler2023diffsep} or separately trained
priors for constituent communication signals
~\citep{jayashankar2023scorebased}. 
These approaches rely on training data that represent the
possible constituent source and corruption classes. In contrast, we assume a generative prior
only for the desired clean signal; the localized corruption may come from an
unknown class, may vary across observations, and need not be anticipated
during training.

\vspace{-7px}
\paragraph{Localized anomaly and OOD detection.}

Localized anomaly and out-of-distribution (OOD) detection methods aim to identify
spatial regions that deviate from the distribution of inlier data. A broad
range of methods estimate such localized supports using feature-space deviations
~\citep{defard2021padim,roth2022patchcore,liu2023simplenet} or learned anomaly
representations~\citep{zavrtanik2021draem}. More recently, diffusion models have been used as expressive
representations of the inlier distribution, enabling anomalies to be localized
from discrepancies produced during denoising or reconstruction
~\citep{wyatt2022anoddpm,zhang2023diffad,he2024diad}. These works demonstrate that the support of an anomalous or
out-of-distribution component can often be estimated without explicitly
modeling its distribution; our work assumes access to this corruption support and uses it towards recovering the
underlying clean signal.

\vspace{-7px}

\section{Method}
\label{sec:method}
\vspace{-7px}

We propose $\method{}$, a diffusion-based method for removing localized structured corruptions without modeling their distribution. We assume access only to the support $\Omega$ identifying where or when the corruption occurs, and a diffusion prior pretrained on clean signals. A natural strategy under these assumptions is diffusion inpainting: discard the measurements inside $\Omega$ and generate the missing content conditioned on the uncorrupted observations outside $\Omega$. However, such a strategy also discards potentially useful information. In our setting, the values inside $\Omega$ are not missing; rather, they contain a superposition of the clean signal and an unknown structured corruption. Our key insight is that these measurements can be progressively reused to improve an initial inpainting estimate.

$\method{}$ operationalizes this insight through three main components. First, we use diffusion inpainting as an \emph{initialization}, and subsequently recover information from the measurements inside the corrupted region $\Omega$ through an alternating signal--corruption decomposition. Second, because the residual between the measurement and the current reconstruction can contain both corruption and incorrectly reconstructed clean content, we use variance across multiple diffusion samples to quantify uncertainty in the current reconstruction. This uncertainty determines the scale of the subsequent diffusion-based refinement. Third, we use the same uncertainty estimate to inform an adaptive stopping criterion.
The complete procedure is summarized in \Cref{alg:cloud}, with key steps explained below. 

\begin{algorithm}[t]
\caption{$\method$: Support-Guided OOD Removal with Diffusion}
\label{alg:cloud}
\begin{algorithmic}[1]

\Require Observation $y$, corruption support $\Omega$, corruption mask $M_\Omega$,
clean-signal diffusion denoiser $D_\theta$, inpainting operator $R_\theta$, 
number of diffusion samples $K$, maximum iterations $M$,
$\{\sigma_m\}_{m=1}^{M}$, stopping threshold $\tau_{\mathrm{stop}}$

\Statex \textbf{Notation:}
$D_\theta(z,\sigma)$ denotes generic diffusion denoising of $z$ from noise level $\sigma$, $R_\theta(y,\Omega)$ denotes inpainting conditioned on the clean measurements $y_{\Omega^c}$. Our experiments instantiate $D_\theta$ via DDIM \citep{song2021denoising} and $R_\theta$ via RePaint~\citep{lugmayr2022repaint}.

\For{$k=1,\ldots,K$}
    \State
    $\widehat{x}_{0}^{(k)}
    \gets
    R_{\theta}(y,\Omega)$,
    \Comment{Initialize with diffusion inpainting}
    \State
    $\displaystyle
    \sigma_{0}^\star
    \gets
    \sqrt{
        \frac{1}{|\Omega|}\sum_{i\in\Omega}
            \operatorname{Var}
            \left(
                \{\widehat{x}_{0,i}^{(k)}\}_{k=1}^K
            \right)
    }$
    \Comment{Estimate uncertainty on $\Omega$}
\EndFor

\For{$m=1,\ldots,M$}

    \For{$k=1,\ldots,K$}
        \State
        $n_m^{(k)}
        \gets
        M_{\Omega}\odot
        \left(
        y-
        \widehat{x}_{m-1}^{(k)}\right)
        $
        \Comment{Estimate additive corruption on $\Omega$}
        \State
        $\widetilde{n}_m^{(k)}
        \gets
        D_\theta
        \left(
            n_m^{(k)}+\epsilon_m^{(k)}\sigma_{m-1}^\star,
            \sigma_{m-1}^\star
        \right),
        ~
        \epsilon_m^{(k)}\sim\mathcal{N}(0,I)$
        \Comment{Refine corruption estimate}

        \State
        $\widetilde{x}_{m}^{(k)}$
        $=
        \begin{cases}
        y_i-\widetilde{n}_{m,i}^{(k)}, & i\in\Omega,\\
        y_i, & i\notin\Omega.
        \end{cases}$
        \Comment{Enforce measurement consistency}

        \State
        $\widehat{x}_{m}^{(k)}
        \gets
        M_{\Omega} \odot D_\theta
        \left(
            \widetilde{x}_{m}^{(k)}
            +\sigma_m\eta_m^{(k)},
            \sigma_m
        \right)+(1-M_{\Omega}) \odot y$
        , \Comment{Refine signal estimate}
        \Statex
        \qquad \qquad \qquad \qquad \qquad \qquad \qquad \qquad \qquad \qquad
        $\eta_m^{(k)}\sim\mathcal{N}(0,I)$

    \EndFor
    \State
    $\displaystyle
    \sigma_{m}^\star
    \gets
    \sqrt{
        \frac{1}{|\Omega|}\sum_{i\in\Omega}
            \operatorname{Var}
            \left(
                \{\widehat{x}_{m,i}^{(k)}\}_{k=1}^K
            \right)
    }$\Comment{Estimate uncertainty on $\Omega$}
    \If{
        $\sigma_{m-1}^\star-\sigma_m^\star
        \leq \tau_{\mathrm{stop}}\sigma^\star_{m-1}$}\Comment{Adaptive stopping criterion}

        \State \Return
        $\displaystyle
        \frac{1}{K}
        \sum_{k=1}^{K}
        \widehat{x}_{m-1}^{(k)}$
    \EndIf
\EndFor

\State \Return
$\displaystyle
\frac{1}{K}
\sum_{k=1}^{K}
\widehat{x}_{M}^{(k)}$

\end{algorithmic}
\end{algorithm}

\label{subsec:inpainting}

We assume that the measurement consists of the clean signal $x$ additively corrupted by an arbitrary structured corruption $n$ within a localized region $\Omega$:
\begin{equation}
y = x + M_{\Omega} \odot n,
\label{eq:measurement}
\end{equation}

where $M_{\Omega}$ is a mask that restricts the corruption to its known support $\Omega$. Importantly, we make no formal distributional assumption on $n$ and do not require examples from its corruption class.

A straightforward approach is to ignore the corrupted measurements $y_{\Omega}$ entirely and perform diffusion inpainting to generate content inside $\Omega$ while preserving the trusted observations $y_{\Omega^c}$. We initialize $\method$ by drawing $K$ such inpainting samples $\{\widehat{x}_{0}^{(k)}\}_{k=1}^K$ via RePaint \citep{lugmayr2022repaint}, though $\method$ is fairly robust to this choice of inpainting method.
The resulting samples are consistent with the clean-signal diffusion prior and the uncorrupted measurements $y_{\Omega^c}$, but they need not recover the true content inside $\Omega$. In particular, inpainting treats the corrupted region as completely unobserved even though $y_{\Omega}$ may still contain substantial information about $x_{\Omega}$.

Our first key idea is to iteratively reintroduce this information after inpainting. At iteration $m$, given the current signal estimate $\widehat{x}_{m-1}^{(k)}$, we form the estimated residual
\begin{equation}
n_m^{(k)}
=
M_{\Omega}\odot
\left(
y-\widehat{x}_{m-1}^{(k)}
\right).
\label{eq:residual}
\end{equation}
If the current signal estimate $\widehat{x}_{m-1}^{(k)}$ were exact, this residual $n_m^{(k)}$ would equal the true corruption $n$. In practice, however, $\widehat{x}_{m-1}^{(k)}$ is imperfect, and consequently $n_m^{(k)}$ can contain a mixture of the true corruption and errors in the estimated clean signal. 
We therefore refine the raw residual before using it to update the clean
signal. Let $D_{\theta}(z,\sigma)$ denote a generic diffusion denoising process starting from noise level $\sigma$.
In our experiments, we instantiate $D_{\theta}$ using the DDIM sampler~\citep{song2021denoising} initialized at noise level $\sigma$ ,
though $\method$ is fairly robust to the choice of diffusion denoiser.
We apply this diffusion refinement operator to each estimated corruption $n_m^{(k)}$:
\begin{equation}
\widetilde{n}_m^{(k)}
=
D_\theta
\left(
n_m^{(k)} + \sigma_{m-1}^\star \epsilon_m^{(k)},
\sigma_{m-1}^\star
\right), 
\qquad
\epsilon_m^{(k)}\sim\mathcal{N}(0,I),
\label{eq:noise_refinement}
\end{equation}
using refinement scale $\sigma_{m-1}^\star$. This scale is adaptive:
\begin{equation}
\sigma_{m-1}^\star
=
\sqrt{
\frac{1}{|\Omega|}\sum_{i\in\Omega}
\operatorname{Var}
            \left(
                \{\widehat{x}_{m-1,i}^{(k)}\}_{k=1}^K
            \right)
},
\label{eq:uncertainty}
\end{equation}
and quantifies how much uncertainty remains in the current signal estimation and corresponding noise estimation. 
This diffusion refinement step helps to resolve fine-scale errors in the estimated corruption, even though it uses a diffusion model trained on the signal class rather than the corruption class. We evaluate this refinement step via an ablation study in \Cref{tab:ablation_nstep}.

\label{subsec:uncertainty}

We then use the refined residual $\widetilde{n}_m^{(k)}$ in conjunction with the measurements $y$ and corruption support $\Omega$ to update our estimate of the clean signal by enforcing measurement consistency:
\begin{equation}
\widetilde{x}_{m}^{(k)}
=
\begin{cases}
y_i-\widetilde{n}_{m,i}^{(k)}, & i\in\Omega,\\
y_i, & i\notin\Omega.
\end{cases}
\label{eq:measurement-update}
\end{equation}

This update ensures
consistency with $y$ but does not guarantee that the resulting signal estimate $\widetilde{x}_{m}^{(k)}$ lies
in the clean-signal distribution captured by the diffusion prior. We therefore alternate the measurement
update with a diffusion-prior refinement following a fixed noise schedule $\sigma_m$:
\begin{equation}
\widehat{x}_{m}^{(k)}=
M_{\Omega} \odot D_\theta
        \left(
            \widetilde{x}_{m}^{(k)}
            +\sigma_m\eta_m^{(k)},
            \sigma_m
        \right)+(1-M_{\Omega}) \odot y
        ,
        ~~
        \eta_m^{(k)}\sim\mathcal{N}(0,I)
\label{eq:diffusion-update}
\end{equation}
$\method$ yields stable performance across a wide range of noise schedules $\sigma_m$; we provide a sensitivity study in \Cref{tab:sigma_schedule_ablation}.

$\method$ alternates between two complementary sources of information: the \emph{measurement}, which reveals what remains unexplained by the current estimates of the signal and the corruption, and the \emph{clean diffusion prior}, which constrains the reconstructed signal toward plausible clean signals and refines details in the estimated corruption. Iterating these updates allows information initially discarded by inpainting to be progressively recovered.

$\method$ terminates adaptively once the uncertainty estimate $\sigma^\star$ stops decreasing much.
When this condition is satisfied, independent clean-signal hypotheses agree sufficiently within the corrupted support, and $\method$ returns their mean.
$\method$ continues its alternating updates until either this condition is satisfied or it reaches a maximum of $M$ iterations.

\vspace{-10px}

\section{Experiments}
\label{sec:experiments}
\vspace{-5px}
We evaluate $\method{}$ on localized structured corruption removal tasks in the
image and audio domains. For images, we consider additive localized structured
corruptions, visible-watermark removal, and shadow removal; for audio, we
consider temporally localized additive interference removal. We report peak
signal-to-noise ratio (PSNR) over the corrupted support $\Omega$ for image
experiments, since full-image metrics can be dominated by unchanged pixels
outside the corrupted region, and scale-invariant signal-to-distortion ratio
improvement (SI-SDRi) within the interference window for audio experiments. Representative results are shown in~\cref{fig:main}.
\vspace{-5px}
\subsection{Additive Localized Structured Image Corruptions}
\label{sec:localized}
\begin{figure}
    \centering
    \vspace{-6pt}
    \includegraphics[width=1\linewidth]{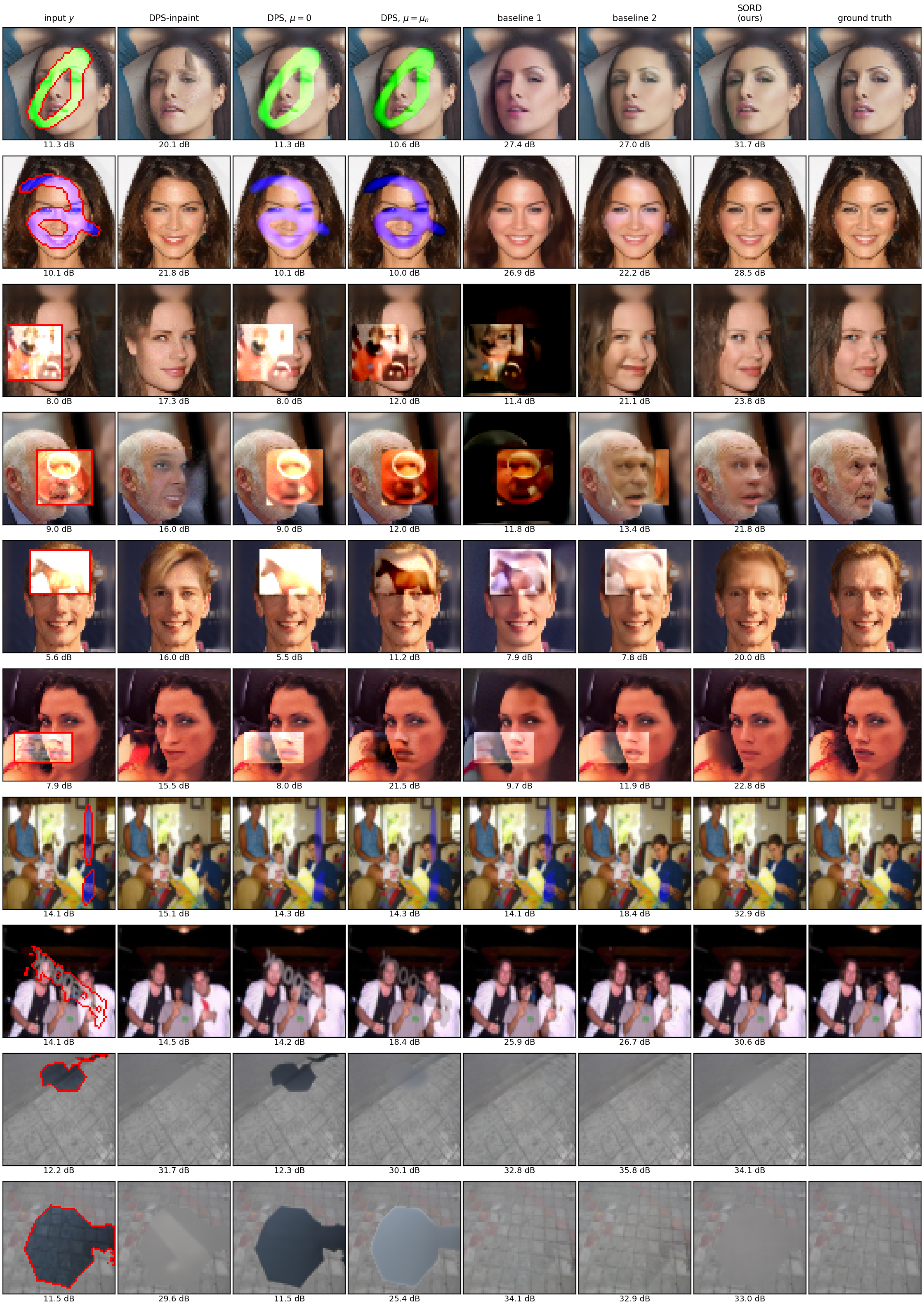}
    \vspace{-20pt}
    
    \caption{
    \textbf{Visual results on localized image corruption removal.}
    Each pair of rows corresponds to a different restoration setting:
    Rows 1--6 respectively show localized MNIST-digit, natural-image patch, and CIFAR-10 patch corruptions on CelebA faces. For those settings, Baseline 1 is \citet{stevens2025removing}, using separate
    CelebA and corruption-patch diffusion priors, while Baseline 2 is its
    mask-projected variant.
    Rows 7--8 show watermark removal on CLWD, with
    PatchWiper~\citep{mo2025patchwiper} as Baseline 1 and
    SLBR~\citep{liang2021slbr} as Baseline 2. 
    Rows 9--10 show shadow removal on ISTD, with
    ShadowDiffusion~\citep{guo2023shadowdiffusion} as Baseline 1 and
    ShadowFormer~\citep{guo2023shadowformer} as Baseline 2.
    Red contours or bounding boxes indicate the known corruption support $\Omega$.
    Numbers below the reconstructions denote PSNR computed only over $\Omega$.
    }
    \label{fig:main}
\end{figure}

\vspace{-5px}
We first consider additive localized corruptions on CelebA
images~\citep{liu2015faceattributes}, constructed by adding natural-image patches~\citep{vanderwalt2014scikit},
MNIST digits~\citep{lecun1998gradient}, and CIFAR-10
patches~\citep{krizhevsky2009learning}. These experiments test the target
setting of this work: the corruption is additive, localized, and highly
structured, while the clean diffusion prior is trained only on CelebA. Visual and quantitative results are reported in \Cref{fig:main} and \Cref{tab:localized_main}, respectively. 

We compare against several DPS~\citep{chung2023dps} baselines using the same CelebA clean prior.
\emph{DPS-Inpaint} treats the corrupted support as missing and inpaints it by conditioning the sampling process on
the observations outside $\Omega$. We additionally evaluate DPS under a
Gaussian likelihood assumption within $\Omega$, either using the oracle corruption mean
$\mu$ or fixing $\mu=0$, since in practice the corruption mean is inaccessible. The limited performance of these variants suggests that the evaluated Gaussian likelihood approximations are inadequate
to remove highly structured localized corruptions. We also compare against SGM \citep{stevens2025removing}, which uses separate diffusion models for the CelebA image distribution and each corruption distribution, as well as a variant of SGM which we endow with additional knowledge of the corruption support $\Omega$. $\method$ outperforms both variants without requiring a learned corruption-specific prior. Further experimental details are provided in \cref{sec:abaltion-details}.

\subsection{Additive Localized Audio Interference}
\label{sec:audio}

To evaluate whether the same support-guided inference principle extends
beyond images to localized audio interference, we use the
Slakh2100~\citep{manilow2019cutting} dataset of instrumental music to construct additive mixtures in which an
interfering source is present only within a known temporal window
$\Omega$. We compare $\method$ with Multi-Source Diffusion Models
(MSDM)~\citep{mariani2024msdm}. MSDM performs source separation using multiple
source-specific diffusion priors, whereas $\method$ uses only a diffusion prior
trained on the target source together with the temporal support $\Omega$ of the
interference.  The audio mixture is constructed as \begin{equation} y=x+g\cdot\left({M}_{\Omega}\odot s\right), \end{equation} where $x$ is the target source and $s$ is the interfering source. We adjust $g$ to control the interference-window Signal to Interference Ratio, SIR, levels between $0$ and $5$ dB, and consider the default Slakh mixing level $g=1$ without SIR-based gain normalization.

We consider both \emph{in-vocabulary} (IV) and
\emph{out-of-vocabulary} (OOV) interfering sources, relative to the MSDM baseline. For a given target source among the MSDM-modeled sources \{piano,
guitar, bass, drums\}, the IV set consists of the other three MSDM-modeled
sources, while the OOV set consists of the remaining eight Slakh instrument
classes for which MSDM has no corresponding diffusion model. 
We additionally consider human speech interference both as an OOV source for the original MSDM model and as an IV source for an augmented version of MSDM endowed with a speech diffusion model.
Quantitative results are presented in \Cref{tab:localized_main}, with per-source results in \Cref{tab:localized_extended} and representative spectrogram visualizations in \Cref{fig:audio_extended} in the appendix. We find that MSDM can outperform $\method$ for certain target instruments and when the interferer is a modeled source (IV), but that $\method$ is generally more robust to unexpected (OOV) interferer types as it does not rely on prior knowledge of the corruption class.

\vspace{-7px}
\subsection{Watermark Removal}
\label{sec:watermark}
\vspace{-7px}
We next consider visible watermarks, which provide a realistic localized
structured corruption containing text or logos with varying color, shape,
scale, and transparency. We use watermark examples from the Colored
Large-scale Watermark Dataset (CLWD)~\citep{liu2021wdnet}.

A watermarked observation follows the alpha-compositing model
\begin{equation}
    y = (1-\alpha)\odot x + \alpha\odot w,
    \label{eq:watermark_model}
\end{equation}
where $x$ is the clean image, $w$ is the watermark appearance, and
$\alpha$ is its opacity map, which may vary at the pixel level. We define the corruption support as all pixels
with nonzero watermark opacity. $\method$ and DPS use only
a diffusion prior for the clean image. In contrast, supervised watermark-removal baselines PatchWiper \citep{mo2025patchwiper} and SLBR \citep{liang2021slbr}
are trained directly on paired watermarked and clean images. 

Although we assume knowledge of the alpha-compositing form in
\Cref{eq:watermark_model}, we do not provide the true pixel-wise opacity map
$\alpha$ to $\method$ at inference time. This deliberately introduces model mismatch, since the
true watermark opacity varies spatially across the logo.

\newcommand{\mstd}[2]{%
    \makebox[2.7em][r]{\ensuremath{#1}}\,
    {\scriptsize$\pm\,#2$}%
}
\newcommand{\bmstd}[2]{%
    \makebox[2.7em][r]{\ensuremath{\mathbf{#1}}}\,
    {\scriptsize$\pm\,#2$}%
}

\begin{table*}[t]
    \centering
    \caption{\label{tab:localized_main}
    \textbf{Localized structured corruption removal across image and audio domains.}
    Image results report PSNR (dB) within $\Omega$ on $100$ CelebA images at
    $64\times64$ resolution.
    Audio results report SI-SDRi improvement on $\Omega$, and the target is piano for all of the audio experiments.
    All values are mean $\pm$ standard deviation.
    IV and OOV denote in-vocabulary and out-of-vocabulary interferers for MSDM,
    respectively; for speech, IV uses MSDM augmented with a speech prior,
    while OOV uses the original four-source MSDM without a speech prior.
    The setting $g=1$ denotes the unnormalized mixture.
    SGM and MSDM use explicit contaminant/source priors, whereas
    $\method$ uses only the clean source prior and corruption support $\Omega$. DPS $\mu=\mu_n$ uses the oracle per-image mean of the true corruption on its support.
    The highest mean in each row is shown in \textbf{bold}. $N$ denotes the number of evaluated mixtures.
    }

    \setlength{\tabcolsep}{3.0pt}
    \renewcommand{\arraystretch}{1.12}

    \resizebox{\textwidth}{!}{%
    \begin{tabular}{@{}lccccccc@{}}

        \toprule
        \textbf{Image Corruption}
        & \textbf{Input}
        & \shortstack{\textbf{DPS}\\\textbf{Inpaint}}
        & \shortstack{\textbf{DPS}\\$\boldsymbol{\mu=\mu_n}$}
        & \shortstack{\textbf{DPS}\\$\boldsymbol{\mu=0}$}
        & \shortstack{\textbf{SGM}\\\textbf{2 priors}}
        & \shortstack{\textbf{SGM}\\\textbf{2 priors + mask}}
        & $\textbf{\method}$
        \\
        \midrule

        MNIST
        & \mstd{12.09}{2.47}
        & \mstd{19.85}{2.77}
        & \mstd{11.06}{1.18}
        & \mstd{11.91}{2.17}
        & \mstd{25.78}{3.42}
        & \mstd{22.51}{3.89}
        & \bmstd{27.16}{2.04}
        \\

        Natural patch
        & \mstd{9.37}{1.62}
        & \mstd{16.84}{2.94}
        & \mstd{15.74}{2.75}
        & \mstd{9.32}{1.58}
        & \mstd{18.92}{7.15}
        & \mstd{19.38}{6.38}
        & \bmstd{21.96}{2.32}
        \\

        CIFAR-10 patch
        & \mstd{7.92}{1.90}
        & \mstd{17.28}{3.92}
        & \mstd{14.87}{2.18}
        & \mstd{7.83}{1.89}
        & \mstd{14.36}{3.93}
        & \mstd{15.34}{4.16}
        & \bmstd{22.05}{3.44}
        \\

        \midrule
        \addlinespace[2pt]


        \textbf{Audio Interferer}
        & \textbf{Mix}
        & $\boldsymbol{N}$
        & \shortstack{\textbf{DPS}\\\textbf{Inpaint}}
        & \shortstack{\textbf{DPS}\\$\boldsymbol{\mu=\mu_n}$}
        & \shortstack{\textbf{DPS}\\$\boldsymbol{\mu=0}$}
        & \textbf{MSDM}
        & $\textbf{\method}$
        \\
        \midrule

        IV instrument
        & SIR $0$
        & $60$
        & \mstd{-0.34}{6.17}
        & \mstd{-0.01}{0.05}
        & \mstd{-0.06}{0.43}
        & \bmstd{5.32}{7.64}
        & \mstd{1.12}{5.11}
        \\

        IV instrument
        & SIR $5$
        & $60$
        & \mstd{-0.12}{5.67}
        & \mstd{0.00}{0.07}
        & \mstd{0.01}{0.09}
        & \mstd{1.40}{6.85}
        & \bmstd{2.74}{6.23}
        \\

        IV instrument
        & $g=1$
        & $60$
        & \mstd{0.34}{0.92}
        & \mstd{-0.02}{0.13}
        & \mstd{-0.02}{0.12}
        & \bmstd{6.48}{9.06}
        & \mstd{1.29}{2.58}
        \\

        Speech (IV)
        & SIR $0$
        & $160$
        & \mstd{1.02}{1.18}
        & \mstd{-0.01}{0.10}
        & \mstd{0.00}{0.05}
        & \bmstd{8.25}{8.64}
        & \mstd{5.51}{3.37}
        \\

        Speech (IV)
        & SIR $5$
        & $160$
        & \mstd{1.85}{2.02}
        & \mstd{-0.03}{0.28}
        & \mstd{0.00}{0.05}
        & \mstd{2.00}{10.03}
        & \bmstd{6.40}{3.39}
        \\

        Speech (IV)
        & $g=1$
        & $160$
        & \mstd{0.94}{1.42}
        & \mstd{0.00}{0.02}
        & \mstd{0.00}{0.02}
        & \bmstd{9.74}{9.73}
        & \mstd{3.74}{3.33}
        \\

        \addlinespace[2pt]


        OOV instrument
        & SIR $0$
        & $160$
        & \mstd{-0.43}{17.01}
        & \mstd{-1.14}{14.35}
        & \mstd{-1.22}{15.50}
        & \mstd{-0.24}{16.94}
        & \bmstd{1.15}{17.54}
        \\

        OOV instrument
        & SIR $5$
        & $160$
        & \mstd{0.03}{17.15}
        & \mstd{-1.21}{15.50}
        & \mstd{-1.13}{14.35}
        & \mstd{-2.67}{18.39}
        & \bmstd{1.97}{17.54}
        \\

        OOV instrument
        & $g=1$
        & $160$
        & \mstd{-0.34}{17.08}
        & \mstd{-1.23}{15.50}
        & \mstd{-1.23}{15.50}
        & \mstd{-1.54}{17.40}
        & \bmstd{0.79}{17.41}
        \\

        Speech (OOV)
        & SIR $0$
        & $105$
        & \mstd{1.14}{1.35}
        & \mstd{-0.01}{0.12}
        & \mstd{0.00}{0.06}
        & \mstd{3.57}{6.01}
        & \bmstd{5.71}{3.60}
        \\

        Speech (OOV)
        & SIR $5$
        & $105$
        & \mstd{2.06}{2.29}
        & \mstd{-0.04}{0.33}
        & \mstd{0.01}{0.04}
        & \mstd{-0.05}{6.80}
        & \bmstd{6.44}{3.40}
        \\

        Speech (OOV)
        & $g=1$
        & $160$
        & \mstd{0.94}{1.42}
        & \mstd{0.00}{0.02}
        & \mstd{0.00}{0.02}
        & \bmstd{4.15}{6.59}
        & \mstd{3.74}{3.33}
        \\
        \bottomrule
    \end{tabular}%
    }
\end{table*}

We evaluate watermark removal both on the CLWD dataset and on synthetically watermarked images from FFHQ \citep{karras2019stylegan} and Imagenette, a ten-class subset of ImageNet \citep{deng2009imagenet}.
For each dataset, we use the PatchWiper model pretrained on the large-scale Pixabay Real-world Watermark Dataset (PRWD) \citep{mo2025patchwiper}. SLBR is trained on paired watermarked and clean images from each dataset, and $\method$ uses a diffusion model trained only on the clean image datasets without any prior knowledge of watermarks.

Quantitative results
are reported in \Cref{tab:watermark_shadow_results}, with representative
examples included in \Cref{fig:main}. 
In this setting where the watermarking model (\Cref{eq:watermark_model}) deviates from the equal-weight additive corruption model assumed by $\method$, $\method$ is competitive with, but does not always outperform, these domain-specific baselines trained on paired clean and corrupted data.

\vspace{-7px}
\subsection{Shadow Removal}
\label{sec:shadow}
\vspace{-7px}

Shadows constitute a qualitatively different corruption from the additive
perturbations considered above, and thus a further model-mismatch stress test of $\method$. We model a shadowed image as a multiplicative corruption:
\begin{equation}
    y = x \odot s,
    \label{eq:shadow_mult}
\end{equation}
where $x$ is the clean image and $s$ is a spatially varying attenuation field with $s<1$ inside the shadow region $\Omega$ and $s=1$ on $\Omega^c$. To apply its
additive corruption-removal procedure, $\method$ applies its measurement-consistency updates in the log-intensity domain, $\log y = \log x + \log s$, but still applies its pretrained clean-image diffusion model in the image domain (to estimates of $x$ and $s$, not their pixel-wise logarithms).

We evaluate shadow removal on the Image Shadow Triplets Dataset
(ISTD)~\citep{wang2018stacked}, which provides shadowed images, corresponding
shadow-free targets, and binary shadow masks. Since the shadow and shadow-free
images are from separate physical camera captures, they exhibit exposure and illumination
differences even outside the annotated shadow region. 
We further evaluate the same shadow-removal task on FFHQ and Imagenette by synthetically applying ISTD shadows to FFHQ and Imagenette images following \Cref{eq:shadow_mult}.
Further details are provided in~\Cref{sec:abaltion-details}.

We compare $\method$ against the same DPS-based restoration baselines and two
task-specific supervised shadow-removal methods:
ShadowDiffusion~\citep{guo2023shadowdiffusion}, which incorporates an explicit conditional
shadow degradation prior within a diffusion framework, and
ShadowFormer~\citep{guo2023shadowformer}, a transformer-based method that uses
global context between shadow and non-shadow regions for restoration.
For each dataset, $\method$ uses only the corresponding clean-image prior and
the shadow support, whereas ShadowDiffusion ans ShadowFormer are trained with paired clean and shadowed examples from each dataset. 

As shown in \Cref{tab:watermark_shadow_results}, the log-domain formulation performs comparably
to the supervised ShadowDiffusion baseline within the shadow support, despite not
using shadow-specific training data, while ShadowFormer has stronger performance.

\begin{table*}[t]
    \centering
    \caption{
    \textbf{Localized structured corruption removal on watermark and shadow benchmarks.}
    Task-specific baselines use dedicated training data or corruption models,
    whereas $\method{}$ uses only the clean-image prior and the known corruption support.
    DPS $\mu=\mu_n$ uses the oracle per-image mean of the true corruption on the support.
    Highest mean in each row is shown in \textbf{bold}.
    }
    \label{tab:watermark_shadow_results}

    \setlength{\tabcolsep}{3.0pt}
    \renewcommand{\arraystretch}{1.12}

    \resizebox{\textwidth}{!}{%
    \begin{tabular}{@{}lccccccc@{}}
        \toprule
        \addlinespace[2pt]
        \textbf{Benchmark}
        & \textbf{Input}
        & \shortstack{\textbf{DPS}\\\textbf{Inpaint}}
        & \shortstack{\textbf{DPS}\\$\boldsymbol{\mu=\mu_n}$}
        & \shortstack{\textbf{DPS}\\$\boldsymbol{\mu=0}$}
        & \textbf{SLBR}
        & \textbf{PatchWiper}
        & \textbf{\method{}}
        \\
        \midrule
        FFHQ
        & $13.49 \pm 3.20$ & $27.13 \pm 3.81$ & $18.80 \pm 5.11$ & $14.35 \pm 3.89$
        & $18.90 \pm 5.10$ & $23.44 \pm 5.59$ & $\mathbf{27.80 \pm 3.50}$ \\
        CLWD
        & $22.55 \pm 5.33$ & $17.14 \pm 4.74$ & $24.26 \pm 5.15$ & $22.73 \pm 5.30$
        & $\mathbf{31.55 \pm 4.47}$ & $31.37 \pm 6.84$ & $25.09 \pm 4.87$ \\
        Imagenette
        & $13.64 \pm 4.30$ & $24.81 \pm 7.56$ & $19.63 \pm 7.18$ & $14.56 \pm 4.61$
        & $21.66 \pm 5.08$ & $21.79 \pm 6.89$ & $\mathbf{26.43 \pm 6.56}$ \\
        \midrule
        \addlinespace[2pt]
        \textbf{Benchmark}
        & \textbf{Input}
        & \shortstack{\textbf{DPS}\\\textbf{Inpaint}}
        & \shortstack{\textbf{DPS}\\$\boldsymbol{\mu=\mu_n}$}
        & \shortstack{\textbf{DPS}\\$\boldsymbol{\mu=0}$}
        & \shortstack{\textbf{Shadow}\\\textbf{Diffusion}}
        & \shortstack{\textbf{Shadow}\\\textbf{Former}}
        & \textbf{\method{}}
        \\
        \midrule
        ISTD
        & $14.48 \pm 3.70$ & $27.44 \pm 7.08$ & $27.60 \pm 4.21$ & $14.57 \pm 3.77$
        & $\mathbf{30.60 \pm 4.93}$ & $30.49 \pm 5.18$ & $29.10 \pm 5.55$ \\
        FFHQ
        & $14.89 \pm 4.13$ & $17.39 \pm 3.85$ & $19.85 \pm 3.80$ & $14.83 \pm 4.09$
        & $19.15 \pm 3.23$ & $\mathbf{23.29 \pm 2.89}$ & $20.34 \pm 4.19$ \\
        Imagenette
        & $14.30 \pm 4.71$ & $15.97 \pm 6.55$ & $19.51 \pm 4.96$ & $14.12 \pm 4.59$
        & $18.57 \pm 2.83$ & $\mathbf{21.79 \pm 4.32}$ & $19.05 \pm 6.50$ \\
        \bottomrule
    \end{tabular}%
    }
\end{table*}

\vspace{-10px}
\subsection{Ablation and Sensitivity Studies}

\vspace{-5px}

We study the sensitivity and robustness of $\method$ to its main hyperparameters and design
choices, with full results in the appendix. Specifically, we examine the effect of the number of samples $K$ in
\Cref{sec:ablation-effect-of-samples}, the noise schedule $\sigma_m$ in
\Cref{tab:sigma_schedule_ablation}, the choice of $D_\theta$ for diffusion
sampling and its computational cost in \Cref{sec:sampling_cost}, the relative value of statistical versus support-based corruption modeling in \Cref{tab:mask_noise_ablation}, and the robustness of $\method{}$ to
imperfect corruption masks in \Cref{sec:mask_ablation}. Finally, because our method refines the
estimated corruption by passing it through the clean-signal diffusion prior at a small
noise level, \Cref{tab:xs_levels} evaluates the effectiveness of this diffusion
residual refinement and its effect on the final reconstruction quality.

\vspace{-10px}

\section{Discussion}
\label{sec:conclusion}
\vspace{-10px}
We introduced $\method$ for removing localized structured corruptions using only a pretrained clean-signal diffusion prior and knowledge of the corruption support, without requiring paired clean and corrupted training data or a separately learned model of the corruption distribution. 
Across image and audio tasks, this support-guided strategy consistently improves over simply discarding the corrupted measurements, as is common in inpainting, and achieves competitive performance against several specialized methods that use corruption-specific training data or priors.

Our experiments also reveal limitations arising from the absence of a corruption-specific prior and the reliance on known corruption support. When the corruption class \emph{is} known and well represented by training data, methods that explicitly model that class can retain an advantage; conversely, $\method$ is particularly useful when the corruption is unexpected or otherwise difficult to model. $\method$ also depends on a sufficiently accurate corruption support: overestimating the support tends to degrade performance gracefully, whereas leaving corrupted pixels outside the support can be substantially more harmful because those observations are treated as trustworthy. Future work may focus on jointly estimating the corruption support alongside corruption removal, developing more principled uncertainty-guided refinement and stopping criteria, and extending the framework to broader non-additive or partially specified corruption models.

\section{Aknowledgement}

This work was supported in part by the HIVES program at Georgia Tech Research Institute (GTRI).

\bibliography{iclr2027_conference}

\begin{thebibliography}{37}
\providecommand{\natexlab}[1]{#1}
\providecommand{\url}[1]{\texttt{#1}}
\expandafter\ifx\csname urlstyle\endcsname\relax
  \providecommand{\doi}[1]{doi: #1}\else
  \providecommand{\doi}{doi: \begingroup \urlstyle{rm}\Url}\fi

\bibitem[Chung et~al.(2023)Chung, Kim, McCann, Klasky, and Ye]{chung2023dps}
Hyungjin Chung, Jeongsol Kim, Michael~T. McCann, Marc~L. Klasky, and Jong~Chul
  Ye.
\newblock Diffusion posterior sampling for general noisy inverse problems.
\newblock In \emph{International Conference on Learning Representations}, 2023.

\bibitem[Cun \& Pun(2021)Cun and Pun]{cun2021split}
Xiaodong Cun and Chi-Man Pun.
\newblock Split then refine: Stacked attention-guided resunets for blind single
  image visible watermark removal.
\newblock In \emph{Proceedings of the AAAI Conference on Artificial
  Intelligence}, volume~35, pp.\  1184--1192, 2021.

\bibitem[Defard et~al.(2021)Defard, Setkov, Loesch, and
  Audigier]{defard2021padim}
Thomas Defard, Aleksandr Setkov, Angelique Loesch, and Romaric Audigier.
\newblock Padim: A patch distribution modeling framework for anomaly detection
  and localization.
\newblock In \emph{International Conference on Pattern Recognition Workshops},
  2021.

\bibitem[Deng et~al.(2009)Deng, Dong, Socher, Li, Li, and
  Fei-Fei]{deng2009imagenet}
Jia Deng, Wei Dong, Richard Socher, Li-Jia Li, Kai Li, and Li~Fei-Fei.
\newblock {ImageNet}: A large-scale hierarchical image database.
\newblock In \emph{Proceedings of the IEEE Conference on Computer Vision and
  Pattern Recognition (CVPR)}, pp.\  248--255, 2009.

\bibitem[Dhariwal \& Nichol(2021)Dhariwal and Nichol]{dhariwal2021diffusion}
Prafulla Dhariwal and Alexander Nichol.
\newblock Diffusion models beat gans on image synthesis.
\newblock In \emph{Advances in Neural Information Processing Systems},
  volume~34, pp.\  8780--8794, 2021.

\bibitem[Guo et~al.(2023{\natexlab{a}})Guo, Huang, Liu, Cheng, and
  Wen]{guo2023shadowformer}
Lanqing Guo, Siyu Huang, Ding Liu, Hao Cheng, and Bihan Wen.
\newblock Shadowformer: Global context helps shadow removal.
\newblock In \emph{Proceedings of the AAAI Conference on Artificial
  Intelligence}, volume~37, pp.\  710--718, 2023{\natexlab{a}}.
\newblock \doi{10.1609/aaai.v37i1.25148}.

\bibitem[Guo et~al.(2023{\natexlab{b}})Guo, Wang, Yang, Huang, Wang, Pfister,
  and Wen]{guo2023shadowdiffusion}
Lanqing Guo, Chong Wang, Wenhan Yang, Siyu Huang, Yufei Wang, Hanspeter
  Pfister, and Bihan Wen.
\newblock Shadowdiffusion: When degradation prior meets diffusion model for
  shadow removal.
\newblock In \emph{Proceedings of the IEEE/CVF Conference on Computer Vision
  and Pattern Recognition}, pp.\  14049--14058, 2023{\natexlab{b}}.

\bibitem[He et~al.(2024)He, Zhang, Chen, Chen, Li, Chen, Wang, Wang, and
  Xie]{he2024diad}
Haoyang He, Jiangning Zhang, Hongxu Chen, Xuhai Chen, Zhishan Li, Xu~Chen,
  Yabiao Wang, Chengjie Wang, and Lei Xie.
\newblock A diffusion-based framework for multi-class anomaly detection.
\newblock In \emph{Proceedings of the AAAI Conference on Artificial
  Intelligence}, volume~38, pp.\  8472--8480, 2024.

\bibitem[Jayashankar et~al.(2023)Jayashankar, Lee, Lancho, Weiss, Polyanskiy,
  and Wornell]{jayashankar2023scorebased}
Tejas Jayashankar, Gary C.~F. Lee, Alejandro Lancho, Amir Weiss, Yury
  Polyanskiy, and Gregory~W. Wornell.
\newblock Score-based source separation with applications to digital
  communication signals.
\newblock In \emph{Advances in Neural Information Processing Systems},
  volume~36, pp.\  5092--5125, 2023.

\bibitem[Karras et~al.(2019)Karras, Laine, and Aila]{karras2019stylegan}
Tero Karras, Samuli Laine, and Timo Aila.
\newblock A style-based generator architecture for generative adversarial
  networks.
\newblock In \emph{Proceedings of the IEEE/CVF Conference on Computer Vision
  and Pattern Recognition (CVPR)}, pp.\  4401--4410, 2019.

\bibitem[Karras et~al.(2024)Karras, Aittala, Lehtinen, Hellsten, Aila, and
  Laine]{karras2024analyzing}
Tero Karras, Miika Aittala, Jaakko Lehtinen, Janne Hellsten, Timo Aila, and
  Samuli Laine.
\newblock Analyzing and improving the training dynamics of diffusion models.
\newblock In \emph{Proceedings of the IEEE/CVF Conference on Computer Vision
  and Pattern Recognition}, pp.\  24174--24184, 2024.

\bibitem[Kawar et~al.(2022)Kawar, Elad, Ermon, and Song]{kawar2022ddrm}
Bahjat Kawar, Michael Elad, Stefano Ermon, and Jiaming Song.
\newblock Denoising diffusion restoration models.
\newblock In \emph{Advances in Neural Information Processing Systems}, 2022.

\bibitem[Kheirandish et~al.(2026)Kheirandish, Hong, and
  Fridovich-Keil]{kheirandish2026klip}
Alireza Kheirandish, Jihoon Hong, and Sara Fridovich-Keil.
\newblock Klip: Localized distribution shift detection via kl-divergence with
  diffusion priors in inverse problems.
\newblock In \emph{Proceedings of the IEEE/CVF Conference on Computer Vision
  and Pattern Recognition}, 2026.

\bibitem[Krizhevsky(2009)]{krizhevsky2009learning}
Alex Krizhevsky.
\newblock Learning multiple layers of features from tiny images.
\newblock Technical report, University of Toronto, 2009.

\bibitem[LeCun et~al.(1998)LeCun, Bottou, Bengio, and
  Haffner]{lecun1998gradient}
Yann LeCun, L{\'e}on Bottou, Yoshua Bengio, and Patrick Haffner.
\newblock Gradient-based learning applied to document recognition.
\newblock \emph{Proceedings of the IEEE}, 86\penalty0 (11):\penalty0
  2278--2324, 1998.

\bibitem[Legin et~al.(2023)Legin, Adam, Hezaveh, and
  Perreault-Levasseur]{legin2023slic}
Ronan Legin, Alexandre Adam, Yashar Hezaveh, and Laurence Perreault-Levasseur.
\newblock Score-based likelihood characterization for inverse problems in the
  presence of non-gaussian noise.
\newblock In \emph{NeurIPS 2023 Workshop on Deep Learning and Inverse
  Problems}, 2023.

\bibitem[Liang et~al.(2021)Liang, Niu, Guo, Long, and Zhang]{liang2021slbr}
Jing Liang, Li~Niu, Fengjun Guo, Teng Long, and Liqing Zhang.
\newblock Visible watermark removal via self-calibrated localization and
  background refinement.
\newblock In \emph{Proceedings of the 29th ACM International Conference on
  Multimedia}, pp.\  4426--4434, 2021.

\bibitem[Liu et~al.(2021)Liu, Zhu, and Bai]{liu2021wdnet}
Yang Liu, Zhen Zhu, and Xiang Bai.
\newblock Wdnet: Watermark-decomposition network for visible watermark removal.
\newblock In \emph{Proceedings of the IEEE/CVF Winter Conference on
  Applications of Computer Vision}, pp.\  3685--3693, 2021.

\bibitem[Liu et~al.(2023)Liu, Zhou, Xu, and Wang]{liu2023simplenet}
Zhikang Liu, Yiming Zhou, Yuansheng Xu, and Zilei Wang.
\newblock Simplenet: A simple network for image anomaly detection and
  localization.
\newblock In \emph{Proceedings of the IEEE/CVF Conference on Computer Vision
  and Pattern Recognition}, pp.\  20402--20411, 2023.

\bibitem[Liu et~al.(2015)Liu, Luo, Wang, and Tang]{liu2015faceattributes}
Ziwei Liu, Ping Luo, Xiaogang Wang, and Xiaoou Tang.
\newblock Deep learning face attributes in the wild.
\newblock In \emph{Proceedings of the IEEE International Conference on Computer
  Vision (ICCV)}, pp.\  3730--3738, 2015.

\bibitem[Lugmayr et~al.(2022)Lugmayr, Danelljan, Romero, Yu, Timofte, and
  Van~Gool]{lugmayr2022repaint}
Andreas Lugmayr, Martin Danelljan, Andres Romero, Fisher Yu, Radu Timofte, and
  Luc Van~Gool.
\newblock {RePaint}: Inpainting using denoising diffusion probabilistic models.
\newblock In \emph{Proceedings of the IEEE/CVF Conference on Computer Vision
  and Pattern Recognition}, pp.\  11461--11471, 2022.

\bibitem[Manilow et~al.(2019)Manilow, Wichern, Seetharaman, and
  Roux]{manilow2019cutting}
Ethan Manilow, Gordon Wichern, Prem Seetharaman, and Jonathan~Le Roux.
\newblock Cutting music source separation some {Slakh}: A dataset to study the
  impact of training data quality and quantity.
\newblock In \emph{Proceedings of the IEEE Workshop on Applications of Signal
  Processing to Audio and Acoustics (WASPAA)}, 2019.

\bibitem[Mariani et~al.(2024)Mariani, Tallini, Postolache, Mancusi, Cosmo, and
  Rodol{\`a}]{mariani2024msdm}
Giorgio Mariani, Irene Tallini, Emilian Postolache, Michele Mancusi, Luca
  Cosmo, and Emanuele Rodol{\`a}.
\newblock Multi-source diffusion models for simultaneous music generation and
  separation.
\newblock In \emph{International Conference on Learning Representations}, 2024.

\bibitem[Mo et~al.(2025)Mo, Chen, Fang, and Li]{mo2025patchwiper}
Zihao Mo, Junye Chen, Chaowei Fang, and Guanbin Li.
\newblock Patchwiper: Leveraging dynamic patch-wise parameters for real-world
  visible watermark removal.
\newblock In \emph{Proceedings of the 33rd ACM International Conference on
  Multimedia}, pp.\  3537--3545. ACM, 2025.
\newblock \doi{10.1145/3746027.3755107}.

\bibitem[Roth et~al.(2022)Roth, Pemula, Zepeda, Sch{\"o}lkopf, Brox, and
  Gehler]{roth2022patchcore}
Karsten Roth, Latha Pemula, Joaquin Zepeda, Bernhard Sch{\"o}lkopf, Thomas
  Brox, and Peter Gehler.
\newblock Towards total recall in industrial anomaly detection.
\newblock In \emph{Proceedings of the IEEE/CVF Conference on Computer Vision
  and Pattern Recognition}, pp.\  14318--14328, 2022.

\bibitem[Scheibler et~al.(2023)Scheibler, Ji, Chung, Byun, Choe, and
  Choi]{scheibler2023diffsep}
Robin Scheibler, Youna Ji, Soo-Whan Chung, Jaeuk Byun, Soyeon Choe, and
  Min-Seok Choi.
\newblock Diffusion-based generative speech source separation.
\newblock In \emph{2023 IEEE International Conference on Acoustics, Speech and
  Signal Processing (ICASSP)}, pp.\  1--5, 2023.
\newblock \doi{10.1109/ICASSP49357.2023.10095310}.

\bibitem[Song et~al.(2021)Song, Meng, and Ermon]{song2021denoising}
Jiaming Song, Chenlin Meng, and Stefano Ermon.
\newblock Denoising diffusion implicit models.
\newblock In \emph{International Conference on Learning Representations}, 2021.

\bibitem[Song et~al.(2023)Song, Vahdat, Mardani, and Kautz]{song2023pigd}
Jiaming Song, Arash Vahdat, Morteza Mardani, and Jan Kautz.
\newblock Pseudoinverse-guided diffusion models for inverse problems.
\newblock In \emph{International Conference on Learning Representations}, 2023.

\bibitem[Stevens et~al.(2025)Stevens, van Gorp, Meral, Shin, Yu, Robert, and
  van Sloun]{stevens2025removing}
Tristan S.~W. Stevens, Hans van Gorp, Faik~C. Meral, Junseob Shin, Jason Yu,
  Jean-Luc Robert, and Ruud J.~G. van Sloun.
\newblock Removing structured noise using diffusion models.
\newblock \emph{Transactions on Machine Learning Research}, 2025\penalty0 (3),
  2025.
\newblock ISSN 2835-8856.
\newblock URL \url{https://openreview.net/forum?id=BvKYsaOVEn}.

\bibitem[van~der Walt et~al.(2014)van~der Walt, Sch{\"o}nberger,
  Nunez-Iglesias, Boulogne, Warner, Yager, Gouillart, Yu, and the scikit-image
  contributors]{vanderwalt2014scikit}
St{\'e}fan van~der Walt, Johannes~L. Sch{\"o}nberger, Juan Nunez-Iglesias,
  Fran{\c{c}}ois Boulogne, Joshua~D. Warner, Neil Yager, Emmanuelle Gouillart,
  Tony Yu, and the scikit-image contributors.
\newblock scikit-image: Image processing in python.
\newblock \emph{PeerJ}, 2:\penalty0 e453, 2014.
\newblock \doi{10.7717/peerj.453}.

\bibitem[Wang et~al.(2018)Wang, Li, and Yang]{wang2018stacked}
Jifeng Wang, Xiang Li, and Jian Yang.
\newblock Stacked conditional generative adversarial networks for jointly
  learning shadow detection and shadow removal.
\newblock In \emph{Proceedings of the IEEE Conference on Computer Vision and
  Pattern Recognition (CVPR)}, 2018.

\bibitem[Wang et~al.(2023)Wang, Yu, and Zhang]{wang2023ddnm}
Yinhuai Wang, Jiwen Yu, and Jian Zhang.
\newblock Zero-shot image restoration using denoising diffusion null-space
  model.
\newblock In \emph{International Conference on Learning Representations}, 2023.

\bibitem[Wyatt et~al.(2022)Wyatt, Leach, Schmon, and
  Willcocks]{wyatt2022anoddpm}
Julian Wyatt, Adam Leach, Sebastian~M. Schmon, and Chris~G. Willcocks.
\newblock {AnoDDPM}: Anomaly detection with denoising diffusion probabilistic
  models using simplex noise.
\newblock In \emph{Proceedings of the IEEE/CVF Conference on Computer Vision
  and Pattern Recognition Workshops}, pp.\  650--656, 2022.

\bibitem[Zavrtanik et~al.(2021)Zavrtanik, Kristan, and
  Sko{\v{c}}aj]{zavrtanik2021draem}
Vitjan Zavrtanik, Matej Kristan, and Danijel Sko{\v{c}}aj.
\newblock Draem: A discriminatively trained reconstruction embedding for
  surface anomaly detection.
\newblock In \emph{Proceedings of the IEEE/CVF International Conference on
  Computer Vision}, pp.\  8330--8339, 2021.

\bibitem[Zhang et~al.(2025)Zhang, Chu, Berner, Meng, Anandkumar, and
  Song]{zhang2025daps}
Bingliang Zhang, Wenda Chu, Julius Berner, Chenlin Meng, Anima Anandkumar, and
  Yang Song.
\newblock Improving diffusion inverse problem solving with decoupled noise
  annealing.
\newblock In \emph{Proceedings of the IEEE/CVF Conference on Computer Vision
  and Pattern Recognition}, pp.\  20895--20905, June 2025.

\bibitem[Zhang et~al.(2023)Zhang, Li, Li, Dai, Jiang, and Xia]{zhang2023diffad}
Xinyi Zhang, Naiqi Li, Jiawei Li, Tao Dai, Yong Jiang, and Shu-Tao Xia.
\newblock Unsupervised surface anomaly detection with diffusion probabilistic
  model.
\newblock In \emph{Proceedings of the IEEE/CVF International Conference on
  Computer Vision}, pp.\  6782--6791, 2023.

\bibitem[Zhu et~al.(2023)Zhu, Zhang, Liang, Cao, Wen, Timofte, and
  Van~Gool]{zhu2023diffpir}
Yuanzhi Zhu, Kai Zhang, Jingyun Liang, Jiezhang Cao, Bihan Wen, Radu Timofte,
  and Luc Van~Gool.
\newblock Denoising diffusion models for plug-and-play image restoration.
\newblock In \emph{Proceedings of the IEEE/CVF Conference on Computer Vision
  and Pattern Recognition Workshops}, pp.\  1219--1229, 2023.

\end{thebibliography}
\bibliographystyle{iclr2027_conference}

\clearpage
\appendix

\section{Appendix}
\label{sec:ablations}

\subsection{Experiment Details}
\label{sec:abaltion-details}
\paragraph{Localized Structured Corruptions:}

All three benchmarks use the same $100$ CelebA test images at $64\times64$ resolution. The natural-image corruptions are $32\times32$ pixel patches and CIFAR-10 corruptions are patches with randomly chosen heights and widths between 16 and 48 pixels.. The corruptions follow the additive model $y = x + M_{\Omega} \odot n$, where $x$ is the clean image and $n$ is the spatially localized corruption. 
For $\method$, we use $K=32$ diffusion samples and a maximum of $M=6$ refinement iterations. We use a noise schedule $\sigma_m$ that starts at $0.4$ and linearly decreases to $0.1$.

We compare against DPS~\citep{chung2023dps} baselines using the same CelebA clean-image diffusion prior.
DPS-Inpaint treats the corrupted support as missing and reconstructs it by conditioning on
the observations outside $\Omega$. We additionally evaluate DPS under a
Gaussian likelihood corruption model assumption within $\Omega$, either using the oracle corruption mean
$\mu=\mu_n$ or fixing $\mu=0$. These variants test whether an explicit Gaussian
noise model is sufficient to remove localized structured corruptions (it usually is not sufficient). We also compare with the two-prior SGM of \citet{stevens2025removing}, which jointly estimates the clean image and corruption using separate diffusion models; CelebA images with MNIST corruption is a main structured-noise setting in that work. For natural image-patch corruptions, we trained the SGM corruption model using synthetic training examples generated by placing randomly selected image patches of size $26\times 26$ to $38\times 38$ on a $64\times64$ black background, separately for each type of image patch corruption. For the masked SGM variant, we enforce the known support by projecting all clean-image estimates at intermediate timesteps to match $y$ on $\Omega^c$.

\paragraph{Localized Audio Interference:}
Each example is a $10$ second clip sampled at $22.05$ kHz, and contains a single target source (piano, in \Cref{tab:localized_main}) together with one interfering source. The interferer is restricted to a single contiguous window covering half of the clip, and this window is provided to all methods.
We also evaluate the same three DPS configurations used for localized image corruptions. DPS uses only the target-source prior and the exact temporal window, with $250$ Karras--Euler reverse steps and $y+\sigma_{\max}\epsilon$ initialization. DPS-Inpaint discards the measurements in the interference window, while the remaining variants use an in-window Gaussian noise assumption with either oracle $\mu=\mu_n$ or default $\mu=0$. For $\method$, we use the exact interference window and $K=64$ diffusion samples for the main experiments. 

\paragraph{Watermark Removal:}
CLWD is a paired synthetic benchmark containing
$60,000$ training images generated from $160$ colored watermark designs and
$10,000$ test images generated from $40$ additional watermark designs.
The dataset provides clean background images and their corresponding watermarked
observations, allowing the watermark support to be determined directly.

We consider three watermark-removal benchmarks. For FFHQ, we composite a subset of $12$ CLWD train-split logos onto $25$ held-out FFHQ faces at resolution $256\times256$; for Imagenette, we apply the same $12$ logos to $25$ Imagenette images at resolution $256\times256$. In both cases, the watermark support occupies approximately $3\%$ of the image. For CLWD, we use its test split, containing $40$ test-split logos and $99$ test images at resolution $64\times64$, with an average corruption support of approximately $14\%$ of the image. Composites use the per-pixel watermark opacity specified by the CLWD dataset. 

The clean image priors are off-the-shelf ADM~\citep{dhariwal2021diffusion} checkpoints for FFHQ and ImageNet-256 (for Imagenette). For CLWD, we use an EDM2~\citep{karras2024analyzing} ImageNet-64 model fine-tuned only on the clean images from the CLWD training split. 
For SLBR, we train a domain-matched SLBR model from scratch on paired data from the corresponding training domain: paired FFHQ examples for FFHQ, and paired Imagenette examples for Imagenette. Training uses all of the CLWD train/test watermarks except for the 12 watermarks which we used for testing for FFHQ and Imagenette. We have followed the published training code for reproducing results. PatchWiper uses the published PRWD~\citep{mo2025patchwiper} checkpoint and predicts its own watermark support.
All DPS variants use the same clean prior and exact corruption support as $\method$. DPS-Inpaint discards the observations within $\Omega$, while the two likelihood-based variants use a per-image Gaussian noise assumption with mean set either to the oracle mean of the true corruption within $\Omega$, $\mu=\mu_n$, or to zero as a standard default. We use $250$ reverse sampling steps for resolution $64\times64$ and $1000$ steps for resolution $256\times256$. 

\paragraph{Shadow Removal:}
For the ISTD benchmark, we use $100$ real test shadows at resolution $64\times64$. To isolate restoration inside the declared shadow support, we recompose each example so that the corrupted input exactly matches the shadow-free target outside $\Omega$; the shadow support occupies approximately $17\%$ of the image on average. The clean prior is an EDM2 model fine-tuned only on shadow-free ISTD images.  We additionally evaluate cross-domain transfer by applying $25$ real ISTD shadow fields as multiplicative attenuation to held-out FFHQ and Imagenette images at resolution $256\times256$. These experiments use the corresponding off-the-shelf FFHQ and ImageNet-256 clean-image priors, respectively, with no shadow-specific training on either target domain.

\subsection{Effect of the number of samples}
\label{sec:ablation-effect-of-samples}
The number of parallel samples $K$ controls both the diversity of the initial inpainting hypotheses and the estimation of the ensemble spread
$\sigma^\star$ used during refinement. We evaluate
$K\in\{2,4,8,16,32,64\}$ across the five image restoration settings in
\Cref{tab:ablation_k}. Increasing $K$ generally improves the final
reconstruction quality. Averaged over the five benchmarks, the corrupted-region PSNR of
$\method$ increases from $21.37$ dB at $K=2$ to $24.84$ dB at $K=32$.
The improvement largely saturates beyond this point. We therefore use $K=32$ as the
default throughout our experiments, providing a favorable trade-off between
ensemble stability and computation.

\begin{table}[t]
    \centering
    \caption{
    \textbf{Ablation on the number of diffusion samples $K$.}
    Each entry reports corrupted-region PSNR (dB) of $\method$
     as mean $\pm$ standard deviation over
    $25$ images per benchmark.
    $K=1$ is not reported because $\sigma^\star$ is the spread across samples.
    Best mean in each column is shown in \textbf{bold}.
    }
    \label{tab:ablation_k}

    \setlength{\tabcolsep}{4pt}
    \renewcommand{\arraystretch}{1.13}

    \resizebox{\linewidth}{!}{%
    \begin{tabular}{@{}lccccc@{}}
        \toprule
        $K$
        & \textbf{Natural patch}
        & \textbf{MNIST digit}
        & \textbf{CIFAR-10 patch}
        & \textbf{CLWD watermark}
        & \textbf{ISTD shadow} \\
        \midrule
        $2$
        & \mstd{18.26}{2.37}
        & \mstd{24.30}{2.50}
        & \mstd{17.44}{2.64}
        & \mstd{23.39}{3.72}
        & \mstd{27.45}{6.99} \\

        $4$
        & \mstd{20.42}{2.28}
        & \mstd{25.46}{2.03}
        & \mstd{20.07}{2.96}
        & \mstd{23.63}{3.93}
        & \mstd{28.77}{6.06} \\

        $8$
        & \mstd{21.74}{2.18}
        & \mstd{26.54}{1.81}
        & \mstd{21.36}{3.20}
        & \mstd{24.41}{3.81}
        & \mstd{29.60}{6.19} \\

        $16$
        & \mstd{22.22}{2.08}
        & \mstd{27.09}{1.48}
        & \mstd{21.91}{3.54}
        & \mstd{24.68}{3.87}
        & \mstd{30.08}{5.99} \\

        $\mathbf{32}$ (default)
        & \mstd{22.52}{2.03}
        & \mstd{27.28}{1.43}
        & \mstd{22.23}{3.41}
        & \mstd{25.04}{3.97}
        & \bmstd{30.13}{6.02} \\

        $64$
        & \bmstd{22.54}{1.84}
        & \bmstd{27.40}{1.44}
        & \bmstd{22.41}{3.43}
        & \bmstd{25.06}{4.19}
        & \mstd{29.98}{6.20} \\
        \bottomrule
    \end{tabular}%
    }
\end{table}

\subsection{Sampling Cost}
\label{sec:sampling_cost}
We next examine whether the diffusion denoiser $D_\theta$ in
$\method$, currently instantiated using DDIM~\citep{song2021denoising}, can be replaced by a faster sampler. We replace only the $x$-refinement step while keeping the
inpainting initialization and $n$-refinement fixed.
As shown in \Cref{tab:xs_levels}, reducing the number of steps does not significantly affect the final results. Watermark removal experiments in which the ground-truth opacity map $\alpha$ is used in the measurement model are labeled as ``declared $\alpha$.''

\begin{table*}[t]
    \centering
    \caption{
    \textbf{Number of DDIM steps in the $x$-refinement step.}
    Our main results use the ``shipped" $\method$ variant, which runs the $x$-refinement step following DDIM over every
    ladder level at or below $\sigma_m$ (13 levels at the largest noise level). We test computationally cheaper variants that cap the ladder at $8$ or $4$ DDIM levels, with everything
    else unchanged ($K$-sample inpainting initialization).
    Image rows report corrupted-region PSNR (dB); audio rows report corruption-window SI-SDRi (dB), both
    as mean $\pm$ standard deviation.
    Brackets report the number of network evaluations per image spent in the
    alternating loop ($x$-step and $n$-step)  excluding the
    initialization, which is identical across columns.
    The highest mean in each row is shown in \textbf{bold}.
    }
    \label{tab:xs_levels}

    \setlength{\tabcolsep}{4pt}
    \renewcommand{\arraystretch}{1.13}

    \resizebox{\textwidth}{!}{%
    \begin{tabular}{@{}lrccc@{}}
        \toprule
        \textbf{Bench} & $N$ & \shortstack{\textbf{All levels}\\\textbf{(default)}} & \textbf{8 DDIM levels} & \textbf{4 DDIM levels} \\
        \midrule
        Natural patch on CelebA
        & 100
        & \bmstd{21.96}{2.33} [1640]
        & \bmstd{21.96}{2.33} [1352]
        & \bmstd{21.96}{2.34} [840] \\

        MNIST digit on CelebA
        & 100
        & \bmstd{27.16}{2.05} [1190]
        & \bmstd{27.16}{2.05} [966]
        & \mstd{27.15}{2.04} [614] \\

        CIFAR-10 patch on CelebA
        & 100
        & \bmstd{22.05}{3.45} [1640]
        & \bmstd{22.05}{3.45} [1352]
        & \bmstd{22.05}{3.45} [840] \\

        \midrule
        CLWD $+$ watermark (declared $\alpha$)
        & 99
        & \mstd{28.06}{6.48} [1184]
        & \mstd{28.06}{6.48} [960]
        & \bmstd{28.07}{6.47} [608] \\

        PRWD $+$ watermark (declared $\alpha$)
        & 100
        & \mstd{30.75}{7.73} [1184]
        & \mstd{30.75}{7.73} [960]
        & \bmstd{30.76}{7.72} [608] \\

        FFHQ $+$ watermark (declared $\alpha$)
        & 25
        & \bmstd{30.05}{3.99} [416]
        & \bmstd{30.05}{3.99} [336]
        & \mstd{30.03}{3.98} [208] \\

        FFHQ $+$ watermark, unseen logos
        & 25
        & \bmstd{26.76}{5.03} [416]
        & \mstd{26.75}{5.02} [336]
        & \mstd{26.74}{5.02} [208] \\

        Imagenette $+$ watermark (declared $\alpha$)
        & 25
        & \bmstd{28.01}{7.15} [320]
        & \bmstd{28.01}{7.15} [240]
        & \mstd{27.98}{7.15} [144] \\

        \midrule
        ISTD shadow (mult., gated)
        & 100
        & \bmstd{29.10}{5.55} [417]
        & \mstd{29.07}{5.54} [257]
        & \mstd{29.00}{5.52} [129] \\

        ISTD shadow $+$ FFHQ (mult., gated)
        & 25
        & \bmstd{20.34}{4.19} [418]
        & \mstd{20.33}{4.19} [258]
        & \mstd{20.32}{4.19} [130] \\

        ISTD shadow $+$ Imagenette (mult., gated)
        & 25
        & \bmstd{19.05}{6.50} [645]
        & \bmstd{19.05}{6.50} [485]
        & \mstd{18.48}{5.80} [212] \\

        \midrule
        Slakh piano (SI-SDRi, $K{=}4$, $M{=}4$)
        & 100
        & \bmstd{3.31}{6.75} [416]
        & \mstd{3.22}{6.82} [288]
        & \mstd{3.08}{6.87} [224] \\

        Slakh guitar (SI-SDRi, $K{=}4$, $M{=}4$)
        & 50
        & \bmstd{3.46}{5.14} [416]
        & \mstd{3.27}{5.31} [288]
        & \mstd{2.78}{5.93} [224] \\

        \bottomrule
    \end{tabular}%
    }
\end{table*}

\subsection{Robustness to Imperfect Corruption Masks}
\label{sec:mask_ablation}

Our formulation assumes access to the support $\Omega$ of the localized
corruption. In practice, however, this support may be obtained from an
approximate segmentation and may not coincide exactly with the true
corrupted region. We therefore evaluate the sensitivity of $\method$ to
systematic mask errors by perturbing the ground-truth support through
dilation, erosion, and spatial shifts.

We evaluate the final iterate with $K=32$ samples and report PSNR over the
\emph{true} corrupted region rather than over the perturbed mask.
\Cref{tab:mask_ablation} reports mean $\pm$ standard deviation, with the
intersection-over-union (IoU) between the supplied and true masks shown in
parentheses.

\begin{table*}[t]
    \centering
    \caption{
    \textbf{Robustness to imperfect corruption masks.}
    We perturb the true corruption support using dilation, erosion, and
    spatial shifts and evaluate PSNR (dB) over the true corrupted region.
    Values in parentheses report IoU between the perturbed and true masks.
    The highest mean in each row is shown in \textbf{bold}.
    }
    \label{tab:mask_ablation}

    \setlength{\tabcolsep}{3pt}
    \renewcommand{\arraystretch}{1.13}

    \resizebox{\textwidth}{!}{%
    \begin{tabular}{@{}lccccc@{}}
        \toprule
        \textbf{Benchmark}
        & \textbf{Exact}
        & \textbf{Dilate 3 px}
        & \textbf{Dilate 5 px}
        & \textbf{Erode 1 px}
        & \textbf{Shift 2 px}
        \\
        \midrule

        Natural patch
        & \bmstd{21.96}{2.33}
        & \mstd{19.94}{2.53} $(0.73)$
        & \mstd{18.85}{2.54} $(0.62)$
        & \mstd{13.84}{2.18} $(0.88)$
        & \mstd{14.38}{2.08} $(0.78)$
        \\

        MNIST digit
        & \bmstd{27.16}{2.05}
        & \mstd{24.40}{2.72} $(0.59)$
        & \mstd{23.32}{2.90} $(0.47)$
        & \mstd{17.83}{1.48} $(0.76)$
        & \mstd{15.20}{2.17} $(0.60)$
        \\

        CLWD watermark
        & \mstd{25.09}{4.33}
        & \mstd{23.60}{3.96} $(0.59)$
        & \mstd{22.39}{3.72} $(0.47)$
        & \bmstd{25.47}{4.91} $(0.75)$
        & \mstd{21.14}{4.26} $(0.61)$
        \\

        CIFAR-10 patch
        & \bmstd{22.05}{3.45}
        & \mstd{19.53}{3.44} $(0.69)$
        & \mstd{18.96}{3.47} $(0.57)$
        & \mstd{12.11}{2.62} $(0.85)$
        & \mstd{12.36}{2.70} $(0.74)$
        \\

        \bottomrule
    \end{tabular}%
    }
\end{table*}

\paragraph{Overestimating the support is comparatively benign.}
Mask dilation produces a gradual degradation rather than a failure of the
refinement procedure. 
In this regime, the primary consideration is that additional clean pixels are
incorrectly treated as unknown and must be reconstructed by the diffusion
prior.

\paragraph{Missing corrupted pixels is substantially more harmful.}
In contrast, underestimating $\Omega$ can leave corrupted measurements
outside the supplied support. Since $\method{}$ assumes that
$y_{\Omega^c}$ is reliable and preserves those observations throughout
refinement, such contamination is effectively locked into the reconstruction.
Consequently, eroding the mask by only one pixel can cause a substantial decrease in mean PSNR.

These results indicate an asymmetric sensitivity to mask error:
{over-covering the suspected corruption is substantially safer than
leaving corrupted pixels outside the mask}.

An exception is watermark removal, as CLWD results are less sensitive to small mask erosions. Because of the way the masks were generated, watermark boundaries
often contain pixels with small or near-zero opacity, so removing these
boundary pixels from the support has little effect. Accordingly, the one-pixel eroded mask performs comparably to, and
in this experiment slightly above, the exact mask.

\begin{figure}[t]
    \centering
    \includegraphics[width=0.9\linewidth]{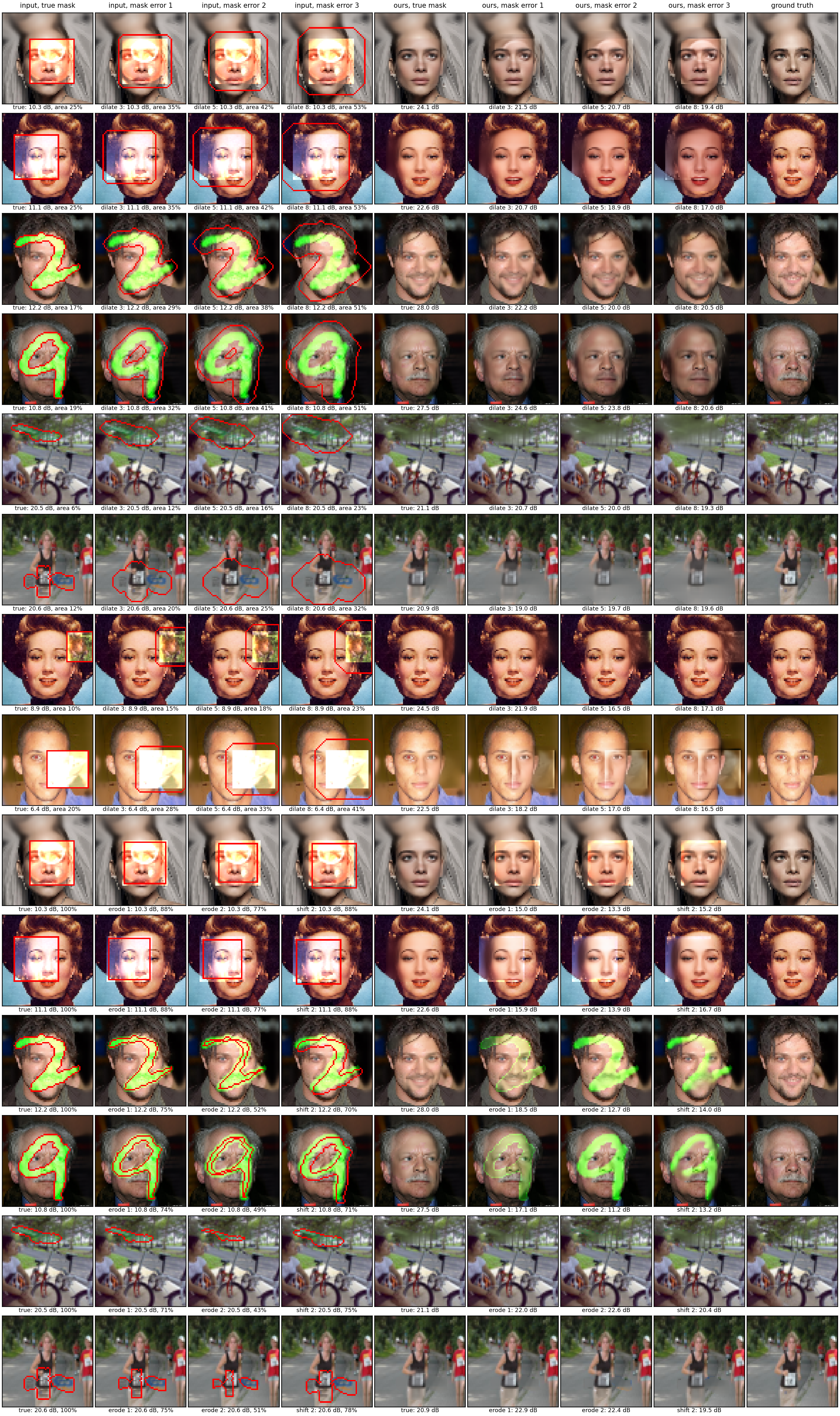}
    \caption{
    \textbf{Qualitative effect of corruption-mask errors.}
    We compare $\method{}$ using the exact corruption support or with dilated,
    eroded, or shifted masks.
    Overestimating the support causes the clean prior to regenerate some
    originally uncorrupted pixels, whereas underestimating the support leaves
    portions of the corruption outside $\Omega$, where they are treated as
    trusted observations and therefore are not corrected.
    }
    \label{fig:mask_error}
\end{figure}

Qualitative examples are shown in \Cref{fig:mask_error}, illustrating that
dilation mainly causes unnecessary regeneration of nearby clean content,
whereas erosion or misalignment leaves visible residual corruption outside the
region that $\method$ is permitted to modify.

\begin{table*}[t]
    \centering
    \caption{
    \textbf{Effect of removing the diffusion $n$-step across benchmarks.}
    The default method re-noises the estimated residual to noise level $\sigma^\star$ and denoises it
    with the clean-signal diffusion prior; ``No DM $n$-step'' instead uses
    the clipped raw residual while keeping all other components unchanged.
    Image results report PSNR (dB) over the corrupted support, while audio
    results report in-window SI-SDR (dB), as mean $\pm$ standard deviation. Audio experiments use $K=4$ and
    $M=4$ for computational efficiency. The highest mean in each column is shown in \textbf{bold}.
    }
    \label{tab:ablation_nstep}

    \setlength{\tabcolsep}{3.5pt}
    \renewcommand{\arraystretch}{1.13}

    \resizebox{\textwidth}{!}{%
    \begin{tabular}{@{}lcccccc@{}}
        \toprule

        \textbf{Method}
        & \shortstack{\textbf{Natural}\\{\scriptsize $N=100$}}
        & \shortstack{\textbf{MNIST}\\{\scriptsize $N=100$}}
        & \shortstack{\textbf{CIFAR-10}\\{\scriptsize $N=100$}}
        & \shortstack{\textbf{CLWD WM}\\{\scriptsize $N=99$}}
        & \shortstack{\textbf{PRWD WM}\\{\scriptsize $N=100$}}
        & \shortstack{\textbf{FFHQ WM}\\{\scriptsize $N=25$}}
        \\
        \midrule

        DM $n$-step
        & \bmstd{21.96}{2.33}
        & \bmstd{27.16}{2.05}
        & \bmstd{22.05}{3.45}
        & \bmstd{24.56}{4.33}
        & \bmstd{23.81}{5.17}
        & \mstd{26.68}{3.58}
        \\

        No DM $n$-step
        & \mstd{20.16}{2.27}
        & \mstd{24.28}{2.36}
        & \mstd{20.40}{3.38}
        & \mstd{22.59}{4.61}
        & \mstd{22.81}{6.32}
        & \bmstd{29.22}{4.08}
        \\

        \midrule
        \addlinespace[3pt]

        \textbf{Method}
        & \shortstack{\textbf{Imagenette WM}\\{\scriptsize $N=25$}}
        & \shortstack{\textbf{ISTD}\\{\scriptsize $N=100$}}
        & \shortstack{\textbf{ISTD on FFHQ}\\{\scriptsize $N=25$}}
        & \shortstack{\textbf{ISTD on Imagenette}\\{\scriptsize $N=25$}}
        & \shortstack{\textbf{Slakh piano}\\{\scriptsize IV+OOV, $N=100$}}
        & \shortstack{\textbf{Slakh guitar}\\{\scriptsize IV+OOV, $N=50$}}
        \\
        \midrule

        DM $n$-step
        & \mstd{24.92}{6.49}
        & \bmstd{29.10}{5.55}
        & \bmstd{20.34}{4.19}
        & \bmstd{19.05}{6.50}
        & \bmstd{3.31}{6.75}
        & \bmstd{3.46}{5.14}
        \\

        No DM $n$-step
        & \bmstd{27.35}{7.41}
        & \mstd{28.83}{5.76}
        & \mstd{19.81}{3.97}
        & \mstd{17.23}{5.49}
        & \mstd{2.85}{7.31}
        & \mstd{2.78}{5.84}
        \\

        \bottomrule
    \end{tabular}%
    }
\end{table*}

\begin{figure}[t]
    \centering
    \includegraphics[
        width=\linewidth,
        height=0.88\textheight,
        keepaspectratio
    ]{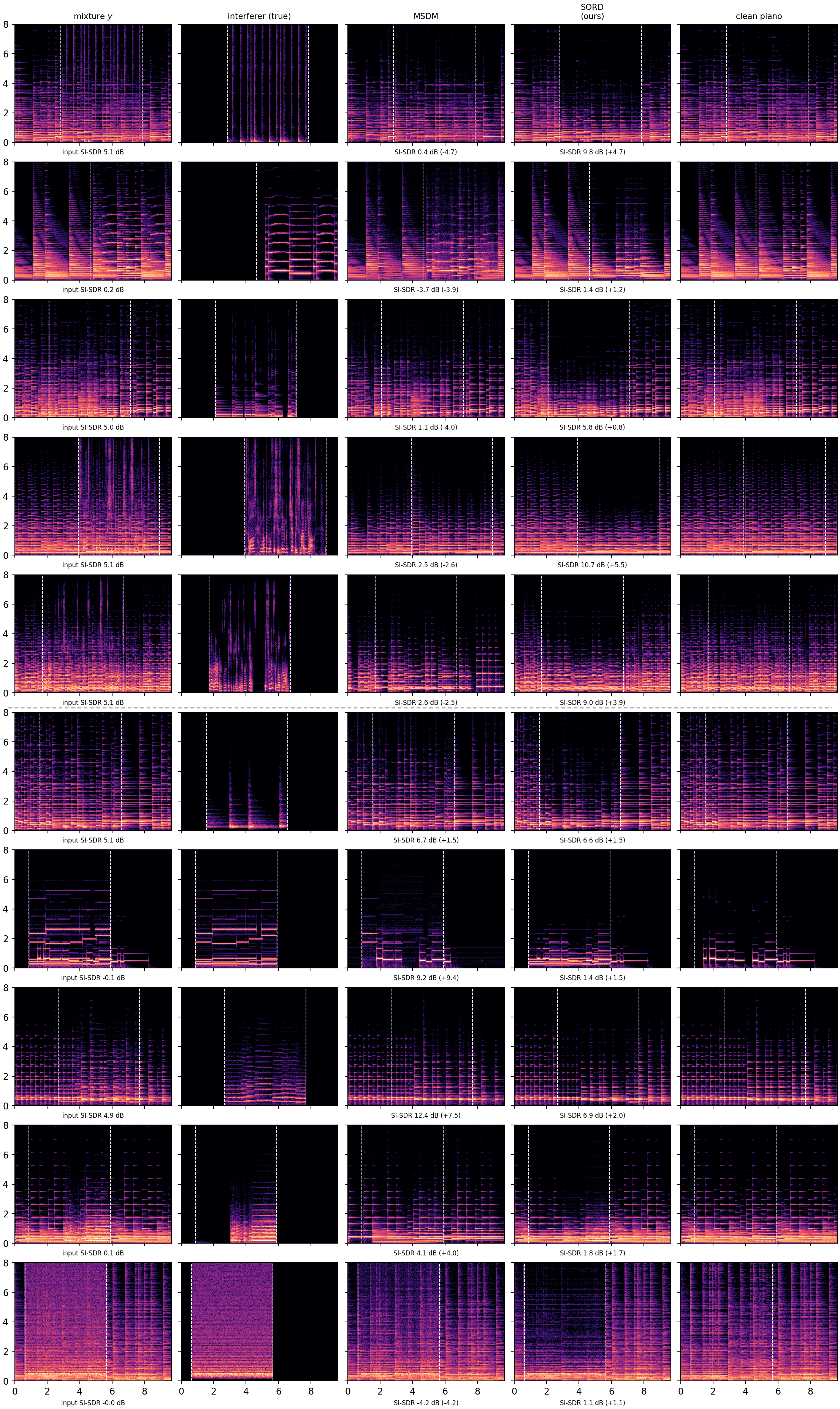}
    \caption{
\textbf{Visualization of localized audio interference removal.}
Each row shows a Slakh piano clip corrupted by a temporally localized
interferer within the dashed interval. Columns show the mixture $y$, the
ground-truth interferer, MSDM, $\method{}$ using only the clean-piano prior,
and the clean target. Rows 1--5 use IV interferers (drums, guitar, bass, and
speech), for which MSDM has corresponding source priors, while Rows 6--10 use
OOV interferers (strings, organ, reed, brass, and synthesizer pad), for which
no matched MSDM prior is available. Values report in-window SI-SDR and, in
parentheses, improvement over the input mixture.}
    \label{fig:audio_extended}
\end{figure}

\begin{table*}[t]
    \centering
    \caption{\textbf{Effect of the re-noising schedule $\sigma_m$.}
    We vary the re-noising schedule across refinement iterations while keeping
    all other components fixed. Results report PSNR (dB) over the corrupted
    support as mean $\pm$ standard deviation. Unless otherwise indicated,
    results are evaluated over 100 images. The highest mean in each column
    is shown in \textbf{bold}.}
    \label{tab:sigma_schedule_ablation}

    \small
    \setlength{\tabcolsep}{5pt}
    \renewcommand{\arraystretch}{1.15}

    \begin{tabular}{@{}lccc@{}}
        \toprule
        \textbf{$\sigma_m$ schedule} & \textbf{Natural} & \textbf{MNIST} & \textbf{CIFAR-10} \\
        \midrule
        $0.4 \rightarrow 0.1$, $M=6$
            & $\mathbf{21.96} \pm 2.33$
            & $\mathbf{27.16} \pm 2.05$
            & $\mathbf{22.05} \pm 3.45$ \\
        Constant $0.8$
            & $20.50 \pm 2.16$
            & $24.41 \pm 1.92\; (n=50)$
            & $20.24 \pm 3.56\; (n=25)$ \\
        $\{0.8,0.4,0.2\}$
            & $20.92 \pm 2.37$
            & $26.21 \pm 2.16$
            & $21.12 \pm 3.49$ \\
        $\{0.9,0.6,0.3,0.1\}$
            & $21.27 \pm 2.38$
            & $26.78 \pm 2.15$
            & $21.43 \pm 3.48$ \\
        $\{0.8,0.4,0.2,0.1\}$
            & $21.34 \pm 2.39$
            & $26.89 \pm 2.15$
            & $21.52 \pm 3.51$ \\
        \bottomrule
    \end{tabular}
\end{table*}

\begin{table*}[t]
    \centering
    \caption{
    \textbf{Effect of support information and contaminant modeling.}
    We ablate the use of the exact corruption support and an additional
    contaminant diffusion model on the three localized CelebA benchmarks.
    Results report regional PSNR (dB) over $\Omega$ as mean $\pm$ standard
    deviation over $100$ images.
    ``No mask'' treats the full image as the possible corruption support,
    whereas ``Exact'' uses the ground-truth support.
    SGM uses the joint $\Pi$GDM sampler of
    \citet{stevens2025removing}.
    The best mean in each column is shown in \textbf{bold}.
    }
    \label{tab:mask_noise_ablation}

    \setlength{\tabcolsep}{3.0pt}
    \renewcommand{\arraystretch}{1.12}

    \resizebox{\textwidth}{!}{%
    \begin{tabular}{@{}lcccccc@{}}
        \toprule

        \textbf{Method}
        & \shortstack{\textbf{MNIST}\\\textbf{No mask}}
        & \shortstack{\textbf{MNIST}\\\textbf{Exact}}
        & \shortstack{\textbf{Natural}\\\textbf{No mask}}
        & \shortstack{\textbf{Natural}\\\textbf{Exact}}
        & \shortstack{\textbf{CIFAR-10}\\\textbf{No mask}}
        & \shortstack{\textbf{CIFAR-10}\\\textbf{Exact}}
        \\
        \midrule

        {\scriptsize Input $y$}
        & \mstd{12.09}{2.49}
        & \mstd{12.09}{2.49}
        & \mstd{9.37}{1.63}
        & \mstd{9.37}{1.63}
        & \mstd{7.92}{1.91}
        & \mstd{7.92}{1.91}
        \\

        \midrule

        {\scriptsize $\method{}$}
        & \mstd{20.29}{3.38}
        & \mstd{27.16}{2.05}
        & \mstd{16.42}{2.69}
        & \mstd{21.96}{2.33}
        & \bmstd{16.00}{3.38}
        & \mstd{22.05}{3.45}
        \\

        {\scriptsize $\method{}$ + matched contaminant DM}
        & \mstd{18.44}{3.68}
        & \bmstd{29.12}{2.26}
        & \mstd{15.54}{2.51}
        & \bmstd{22.98}{2.15}
        & \mstd{14.91}{3.23}
        & \bmstd{22.08}{3.49}
        \\

        \midrule

        {\scriptsize SGM / PiGDM, one prior}
        & \mstd{11.42}{3.11}
        & \mstd{15.60}{2.75}
        & \mstd{13.18}{2.04}
        & \mstd{14.77}{2.45}
        & \mstd{12.33}{2.95}
        & \mstd{13.98}{3.29}
        \\

        {\scriptsize SGM / PiGDM, two priors}
        & \bmstd{25.64}{3.42}
        & \mstd{22.38}{3.89}
        & \bmstd{18.90}{2.11}
        & \mstd{19.42}{3.32}
        & \mstd{14.66}{2.83}
        & \mstd{15.25}{2.77}
        \\

        \bottomrule
    \end{tabular}%
    }
\end{table*}

\begin{table*}[t]
    \centering
    \caption{
    \textbf{Localized structured corruption removal across image and audio domains.}
    Audio results report windowed SI-SDR improvement (SI-SDRi, dB) for
    piano, guitar, bass, and drums targets under localized instrumental and speech interference.
    All values are mean $\pm$ standard deviation.
    IV and OOV denote MSDM in-vocabulary and out-of-vocabulary interferers,
    respectively; for speech, IV uses MSDM augmented with a speech prior,
    while OOV uses the original four-source MSDM without a speech prior.
    The setting $g=1$ denotes the unnormalized mixture.
    SGM and MSDM use explicit contaminant/source priors, whereas
    $\method{}$ uses only the clean target prior.
    The highest mean in each row is shown in \textbf{bold}.
    }
    \label{tab:localized_extended}

    \setlength{\tabcolsep}{3.0pt}
    \renewcommand{\arraystretch}{1.12}

    \resizebox{\textwidth}{!}{%
    \begin{tabular}{@{}lccccccc@{}}
        \toprule
        \textbf{Audio Interferer}
        & \textbf{Mix}
        & $\boldsymbol{N}$
        & \shortstack{\textbf{DPS}\\\textbf{Inpaint}}
        & \shortstack{\textbf{DPS}\\$\boldsymbol{\mu=\mu_n}$}
        & \shortstack{\textbf{DPS}\\$\boldsymbol{\mu=0}$}
        & \textbf{MSDM}
        & \textbf{\method{}}
        \\
        \midrule
        \multicolumn{8}{@{}l}{\textit{Piano target}}\\[-1pt]

        IV instrument
        & SIR $0$ & $60$
        & \mstd{-0.34}{6.17}
        & \mstd{-0.01}{0.05}
        & \mstd{-0.06}{0.43}
        & \bmstd{5.32}{7.64}
        & \mstd{1.12}{5.11}
        \\

        IV instrument
        & SIR $5$ & $60$
        & \mstd{-0.12}{5.67}
        & \mstd{0.00}{0.07}
        & \mstd{0.01}{0.09}
        & \mstd{1.40}{6.85}
        & \bmstd{2.74}{6.23}
        \\

        IV instrument
        & $g=1$ & $60$
        & \mstd{0.34}{0.92}
        & \mstd{-0.02}{0.13}
        & \mstd{-0.02}{0.12}
        & \bmstd{6.48}{9.06}
        & \mstd{1.29}{2.58}
        \\

        Speech (IV)
        & SIR $0$ & $160$
        & \mstd{1.02}{1.18}
        & \mstd{-0.01}{0.10}
        & \mstd{0.00}{0.05}
        & \bmstd{8.25}{8.64}
        & \mstd{5.51}{3.37}
        \\

        Speech (IV)
        & SIR $5$ & $160$
        & \mstd{1.85}{2.02}
        & \mstd{-0.03}{0.28}
        & \mstd{0.00}{0.05}
        & \mstd{2.00}{10.03}
        & \bmstd{6.40}{3.39}
        \\

        \addlinespace[1pt]

        OOV instrument
        & SIR $0$ & $160$
        & \mstd{-0.43}{17.01}
        & \mstd{-1.14}{14.35}
        & \mstd{-1.22}{15.50}
        & \mstd{-0.24}{16.94}
        & \bmstd{1.15}{17.54}
        \\

        OOV instrument
        & SIR $5$ & $160$
        & \mstd{0.03}{17.15}
        & \mstd{-1.21}{15.50}
        & \mstd{-1.13}{14.35}
        & \mstd{-2.67}{18.39}
        & \bmstd{1.97}{17.54}
        \\

        OOV instrument
        & $g=1$ & $160$
        & \mstd{-0.34}{17.08}
        & \mstd{-1.23}{15.50}
        & \mstd{-1.23}{15.50}
        & \mstd{-1.54}{17.40}
        & \bmstd{0.79}{17.41}
        \\

        Speech (OOV)
        & SIR $0$ & $160$
        & \mstd{1.02}{1.18}
        & \mstd{-0.01}{0.10}
        & \mstd{0.00}{0.05}
        & \mstd{3.56}{5.83}
        & \bmstd{5.51}{3.37}
        \\

        Speech (OOV)
        & SIR $5$ & $160$
        & \mstd{1.85}{2.02}
        & \mstd{-0.03}{0.28}
        & \mstd{0.00}{0.05}
        & \mstd{0.11}{6.81}
        & \bmstd{6.40}{3.39}
        \\

        \addlinespace[3pt]
        \cmidrule(lr){1-8}
        \addlinespace[1pt]

        \multicolumn{8}{@{}l}{\textit{Guitar target}}\\[-1pt]

        IV instrument
        & SIR $0$ & $60$
        & \mstd{0.52}{0.85}
        & \mstd{0.00}{0.03}
        & \mstd{0.00}{0.03}
        & \bmstd{4.83}{8.49}
        & \mstd{2.24}{3.02}
        \\

        IV instrument
        & SIR $5$ & $60$
        & \mstd{1.18}{1.69}
        & \mstd{0.01}{0.10}
        & \mstd{0.00}{0.09}
        & \mstd{-0.04}{8.97}
        & \bmstd{4.34}{3.95}
        \\

        IV instrument
        & $g=1$ & $60$
        & \mstd{0.43}{0.74}
        & \mstd{0.00}{0.03}
        & \mstd{0.01}{0.03}
        & \bmstd{5.36}{8.26}
        & \mstd{1.29}{2.68}
        \\

        Speech (IV)
        & SIR $0$ & $110$
        & \mstd{0.62}{4.81}
        & \mstd{-0.21}{2.26}
        & \mstd{-0.22}{2.26}
        & \mstd{3.86}{11.14}
        & \bmstd{4.73}{3.83}
        \\

        Speech (IV)
        & SIR $5$ & $110$
        & \mstd{1.70}{5.29}
        & \mstd{-0.26}{2.74}
        & \mstd{-0.26}{2.75}
        & \mstd{-1.72}{14.26}
        & \bmstd{6.14}{4.83}
        \\

        \addlinespace[1pt]

        OOV instrument
        & SIR $0$ & $160$
        & \mstd{-2.66}{29.39}
        & \mstd{-3.53}{25.66}
        & \mstd{-3.53}{25.68}
        & \mstd{-3.92}{31.61}
        & \bmstd{-0.74}{30.23}
        \\

        OOV instrument
        & SIR $5$ & $160$
        & \mstd{-1.76}{29.59}
        & \mstd{-3.52}{25.68}
        & \mstd{-3.53}{25.68}
        & \mstd{-4.47}{30.81}
        & \bmstd{-0.01}{30.19}
        \\

        OOV instrument
        & $g=1$ & $160$
        & \mstd{-3.39}{29.93}
        & \mstd{-3.89}{26.13}
        & \mstd{-3.86}{25.94}
        & \mstd{-5.88}{32.06}
        & \bmstd{-2.03}{30.70}
        \\

        Speech (OOV)
        & SIR $0$ & $110$
        & \mstd{0.62}{4.81}
        & \mstd{-0.21}{2.26}
        & \mstd{-0.22}{2.26}
        & \mstd{-0.78}{9.12}
        & \bmstd{4.73}{3.83}
        \\

        Speech (OOV)
        & SIR $5$ & $110$
        & \mstd{1.70}{5.29}
        & \mstd{-0.26}{2.74}
        & \mstd{-0.26}{2.75}
        & \mstd{-1.05}{7.60}
        & \bmstd{6.14}{4.83}
        \\

        \addlinespace[3pt]
        \cmidrule(lr){1-8}
        \addlinespace[1pt]

        \multicolumn{8}{@{}l}{\textit{Bass target}}\\[-1pt]

        IV instrument
        & SIR $0$ & $60$
        & \mstd{3.92}{3.57}
        & \mstd{0.00}{0.02}
        & \mstd{0.00}{0.03}
        & \bmstd{21.80}{8.59}
        & \mstd{9.83}{8.02}
        \\

        IV instrument
        & SIR $5$ & $60$
        & \mstd{6.80}{5.88}
        & \mstd{0.00}{0.07}
        & \mstd{0.00}{0.07}
        & \bmstd{19.26}{7.81}
        & \mstd{12.31}{5.96}
        \\

        IV instrument
        & $g=1$ & $60$
        & \mstd{4.64}{4.86}
        & \mstd{0.00}{0.03}
        & \mstd{0.00}{0.03}
        & \bmstd{21.54}{7.98}
        & \mstd{10.06}{7.64}
        \\

        Speech (IV)
        & SIR $0$ & $110$
        & \mstd{4.91}{2.99}
        & \mstd{0.01}{0.02}
        & \mstd{0.01}{0.01}
        & \mstd{8.69}{2.48}
        & \bmstd{12.93}{5.69}
        \\

        Speech (IV)
        & SIR $5$ & $110$
        & \mstd{7.85}{4.40}
        & \mstd{0.00}{0.03}
        & \mstd{0.00}{0.03}
        & \mstd{4.03}{2.59}
        & \bmstd{15.27}{3.80}
        \\

        \addlinespace[1pt]

        OOV instrument
        & SIR $0$ & $160$
        & \mstd{7.45}{5.75}
        & \mstd{0.01}{0.08}
        & \mstd{0.01}{0.09}
        & \bmstd{18.58}{9.82}
        & \mstd{14.32}{7.84}
        \\

        OOV instrument
        & SIR $5$ & $160$
        & \mstd{10.53}{6.14}
        & \mstd{0.00}{0.05}
        & \mstd{0.00}{0.07}
        & \bmstd{16.43}{8.48}
        & \mstd{14.24}{5.19}
        \\

        OOV instrument
        & $g=1$ & $160$
        & \mstd{8.51}{6.15}
        & \mstd{0.02}{0.19}
        & \mstd{0.00}{0.10}
        & \bmstd{17.52}{8.83}
        & \mstd{13.51}{6.92}
        \\

        Speech (OOV)
        & SIR $0$ & $110$
        & \mstd{4.91}{2.99}
        & \mstd{0.01}{0.02}
        & \mstd{0.01}{0.01}
        & \bmstd{17.74}{6.67}
        & \mstd{12.93}{5.69}
        \\

        Speech (OOV)
        & SIR $5$ & $110$
        & \mstd{7.85}{4.40}
        & \mstd{0.00}{0.03}
        & \mstd{0.00}{0.03}
        & \bmstd{15.88}{6.58}
        & \mstd{15.27}{3.80}
        \\

        \addlinespace[3pt]
        \cmidrule(lr){1-8}
        \addlinespace[1pt]

        \multicolumn{8}{@{}l}{\textit{Drums target}}\\[-1pt]

        IV instrument
        & SIR $0$ & $60$
        & \mstd{1.25}{1.00}
        & \mstd{0.01}{0.09}
        & \mstd{0.02}{0.10}
        & \bmstd{19.88}{6.54}
        & \mstd{9.31}{5.48}
        \\

        IV instrument
        & SIR $5$ & $60$
        & \mstd{1.85}{1.36}
        & \mstd{-0.01}{0.13}
        & \mstd{0.00}{0.12}
        & \bmstd{16.54}{7.10}
        & \mstd{9.62}{4.22}
        \\

        IV instrument
        & $g=1$ & $60$
        & \mstd{1.28}{1.09}
        & \mstd{0.00}{0.12}
        & \mstd{0.01}{0.10}
        & \bmstd{19.67}{6.79}
        & \mstd{9.47}{5.55}
        \\

        Speech (IV)
        & SIR $0$ & $110$
        & \mstd{1.32}{0.55}
        & \mstd{0.03}{0.08}
        & \mstd{0.02}{0.07}
        & \bmstd{11.85}{6.41}
        & \mstd{8.89}{3.31}
        \\

        Speech (IV)
        & SIR $5$ & $110$
        & \mstd{1.48}{0.67}
        & \mstd{0.03}{0.10}
        & \mstd{0.01}{0.10}
        & \mstd{7.56}{8.22}
        & \bmstd{10.05}{3.48}
        \\

        \addlinespace[1pt]

        OOV instrument
        & SIR $0$ & $160$
        & \mstd{-0.54}{23.97}
        & \mstd{-2.38}{21.43}
        & \mstd{-2.37}{21.32}
        & \bmstd{12.60}{25.89}
        & \mstd{9.24}{25.33}
        \\

        OOV instrument
        & SIR $5$ & $160$
        & \mstd{0.02}{24.04}
        & \mstd{-2.36}{21.10}
        & \mstd{-2.39}{21.17}
        & \bmstd{9.79}{25.90}
        & \mstd{8.22}{25.12}
        \\

        OOV instrument
        & $g=1$ & $160$
        & \mstd{-0.96}{24.32}
        & \mstd{-2.56}{21.62}
        & \mstd{-2.59}{21.34}
        & \bmstd{10.58}{26.70}
        & \mstd{8.26}{25.77}
        \\

        Speech (OOV)
        & SIR $0$ & $110$
        & \mstd{1.32}{0.55}
        & \mstd{0.03}{0.08}
        & \mstd{0.02}{0.07}
        & \bmstd{9.93}{4.36}
        & \mstd{8.89}{3.31}
        \\

        Speech (OOV)
        & SIR $5$ & $110$
        & \mstd{1.48}{0.67}
        & \mstd{0.03}{0.10}
        & \mstd{0.01}{0.10}
        & \mstd{8.89}{4.98}
        & \bmstd{10.05}{3.48}
        \\

        \bottomrule
    \end{tabular}%
    }
\end{table*}
\end{document}